\documentclass{article}
\usepackage{ijcai22}

\usepackage{times}
\usepackage{soul}
\usepackage{url}
\usepackage[hidelinks]{hyperref}
\usepackage[utf8]{inputenc}
\usepackage[small]{caption}
\usepackage{graphicx}
\usepackage{amsmath}
\usepackage{amsthm}
\usepackage{booktabs}
\usepackage{algorithm}
\usepackage{algorithmic}
\usepackage{multirow}
\usepackage{threeparttable}

\title{DivSwapper: Towards Diversified Patch-based Arbitrary Style Transfer}

\author{
	Zhizhong Wang
	\and
	Lei Zhao\thanks{Corresponding authors.} \and
	Haibo Chen\and
	Zhiwen Zuo\and 
	\\
	Ailin Li \and
	Wei Xing$^*$ \And
	Dongming Lu
	\affiliations
	College of Computer Science and Technology, Zhejiang University
	\emails
	\{endywon, cszhl, cshbchen, zzwcs, liailin, wxing, ldm\}@zju.edu.cn	
}

\begin{document}

\maketitle

\begin{abstract}
  Gram-based and patch-based approaches are two important research lines of style transfer. Recent diversified Gram-based methods have been able to produce multiple and diverse stylized outputs for the same content and style images. However, as another widespread research interest, the diversity of patch-based methods remains challenging due to the stereotyped style swapping process based on nearest patch matching. To resolve this dilemma, in this paper, we dive into the crux of existing patch-based methods and propose a universal and efficient module, termed DivSwapper, for diversified patch-based arbitrary style transfer. The key insight is to use an essential intuition that neural patches with higher activation values could contribute more to diversity. Our DivSwapper is plug-and-play and can be easily integrated into existing patch-based and Gram-based methods to generate diverse results for arbitrary styles. We conduct theoretical analyses and extensive experiments to demonstrate the effectiveness of our method, and compared with state-of-the-art algorithms, it shows superiority in diversity, quality, and efficiency.
\end{abstract}

\renewcommand\arraystretch{1}
\begin{figure*}[t]
	\centering
	\setlength{\tabcolsep}{0.05cm}
	\centering
	\setlength{\tabcolsep}{0.03cm}
	\begin{tabular}{cccccp{0.06cm}|p{0.06cm}ccccc}
		
		\includegraphics[width=0.095\linewidth]{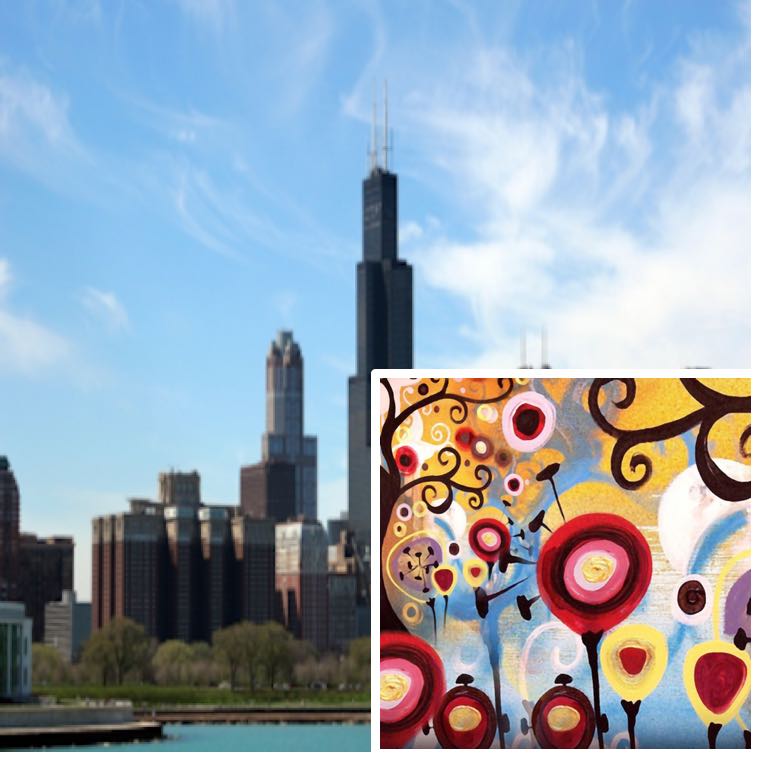}&
		\includegraphics[width=0.095\linewidth]{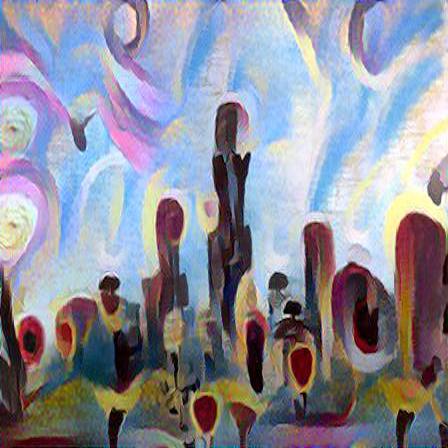}&
		\includegraphics[width=0.095\linewidth]{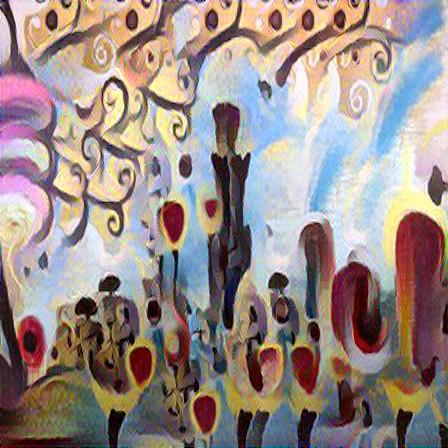}&
		\includegraphics[width=0.095\linewidth]{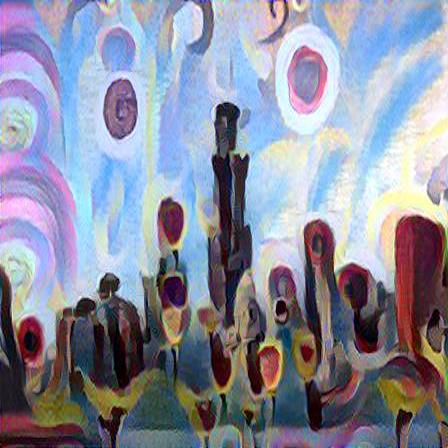}&
		\includegraphics[width=0.095\linewidth]{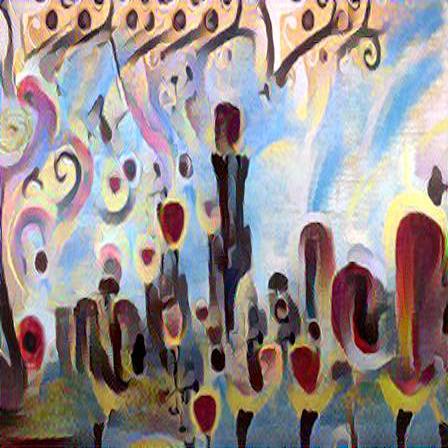}&
		&&
		\includegraphics[width=0.095\linewidth]{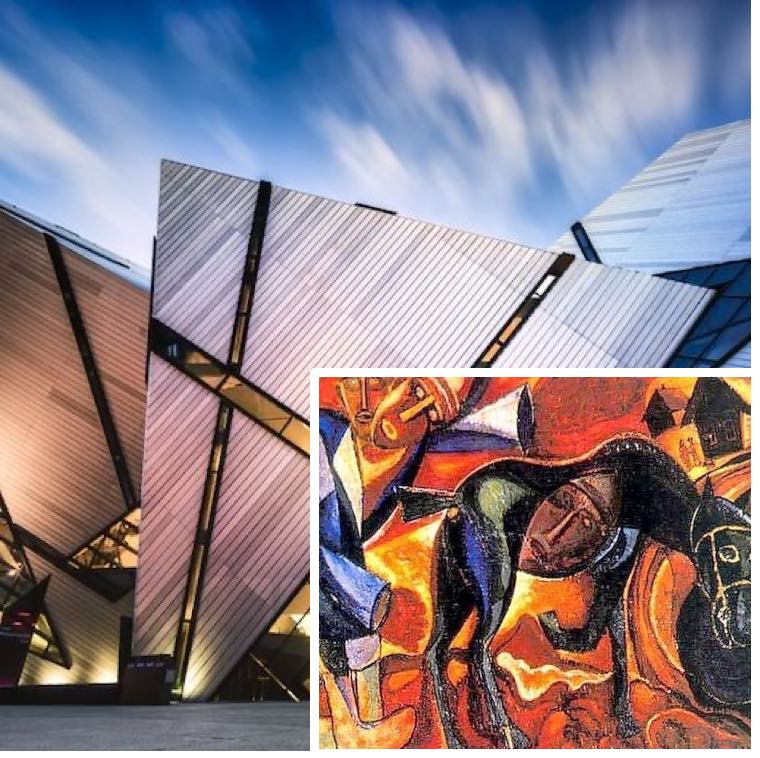}&
		\includegraphics[width=0.095\linewidth]{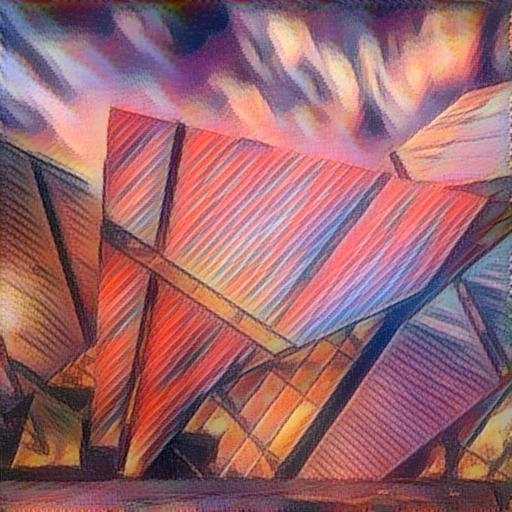}&
		\includegraphics[width=0.095\linewidth]{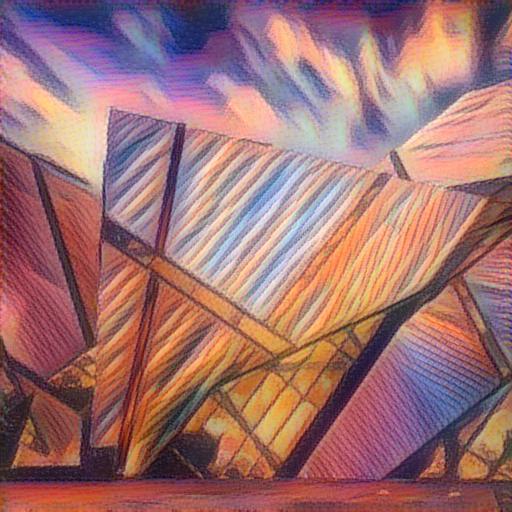}&
		\includegraphics[width=0.095\linewidth]{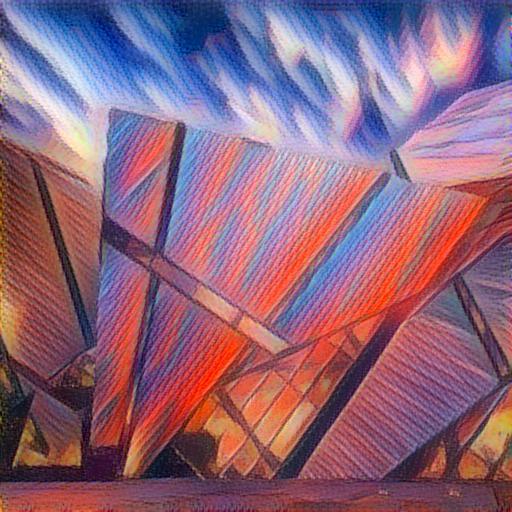}&
		\includegraphics[width=0.095\linewidth]{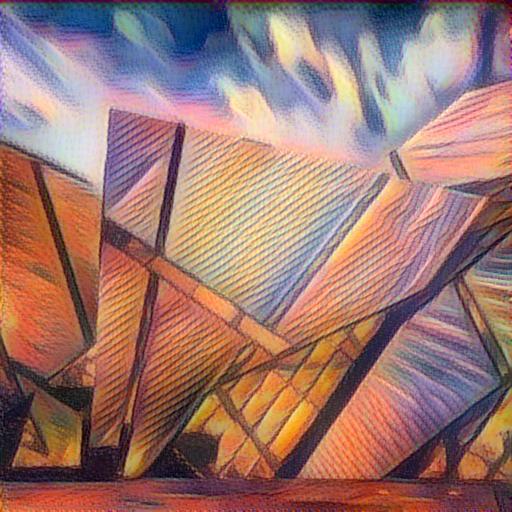}
		\\
		\footnotesize Inputs & \multicolumn{4}{|c}{\footnotesize  (a) CNNMRF + Our DivSwapper}&
		&&
		\footnotesize Inputs & \multicolumn{4}{|c}{\footnotesize  (b) Style-Swap + Our DivSwapper}
		\vspace{0.3em}
		\\
		\includegraphics[width=0.095\linewidth]{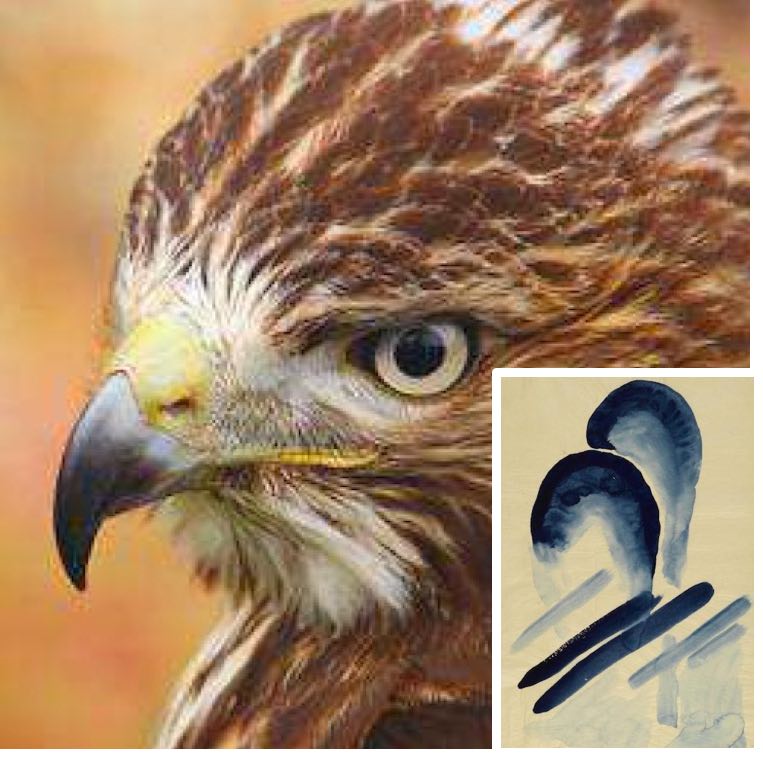}&
		\includegraphics[width=0.095\linewidth]{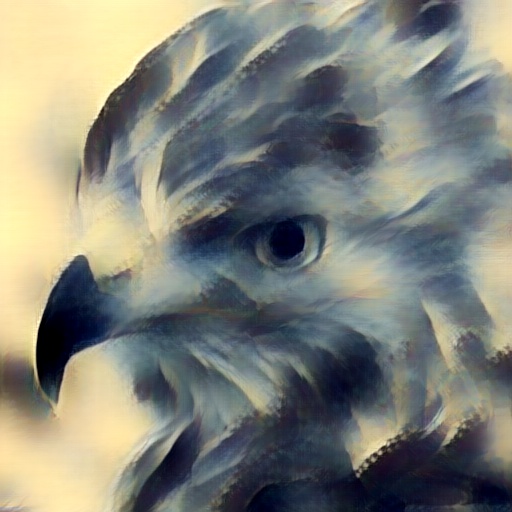}&
		\includegraphics[width=0.095\linewidth]{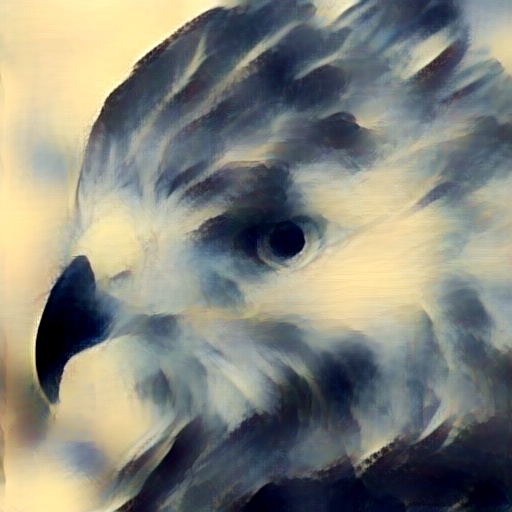}&
		\includegraphics[width=0.095\linewidth]{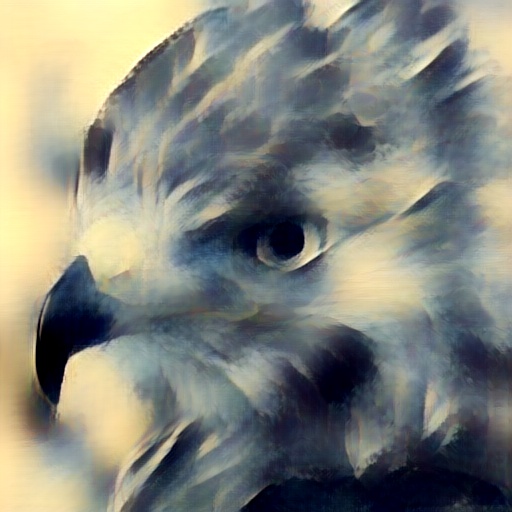}&
		\includegraphics[width=0.095\linewidth]{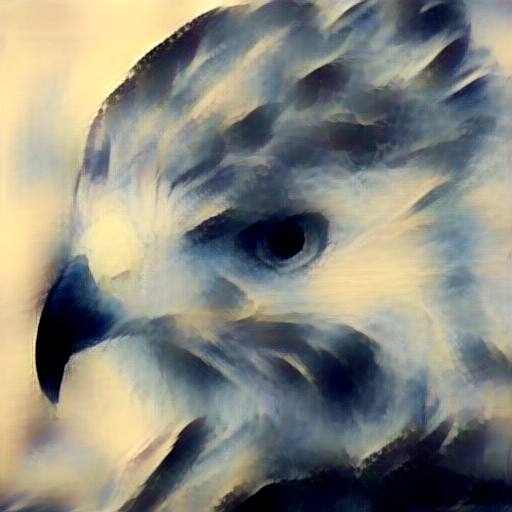}&
		&&
		\includegraphics[width=0.095\linewidth]{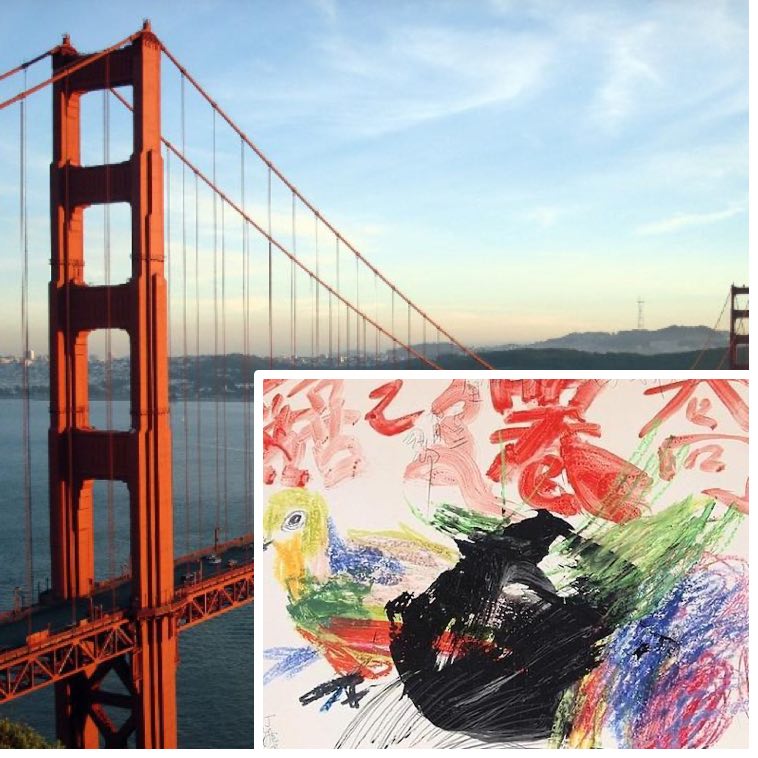}&
		\includegraphics[width=0.095\linewidth]{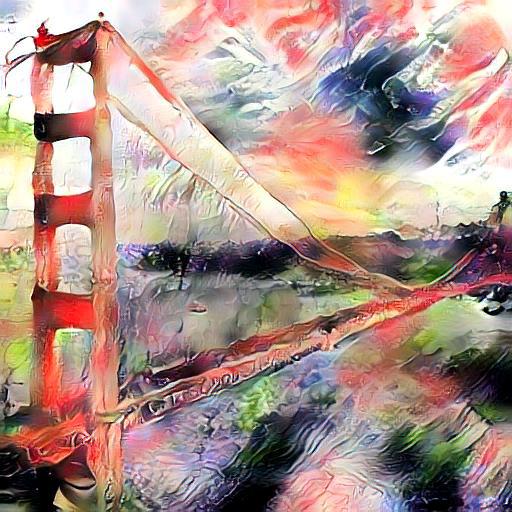}&
		\includegraphics[width=0.095\linewidth]{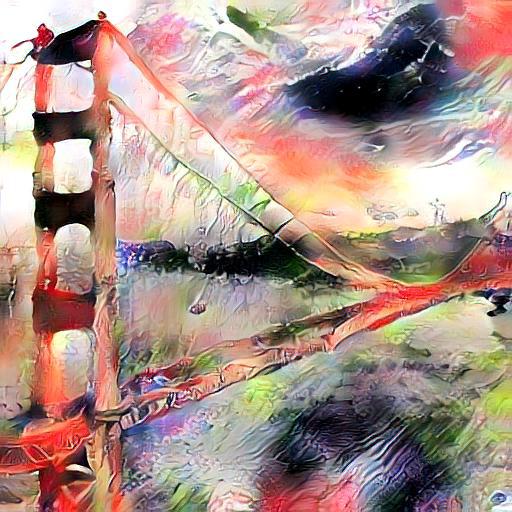}&
		\includegraphics[width=0.095\linewidth]{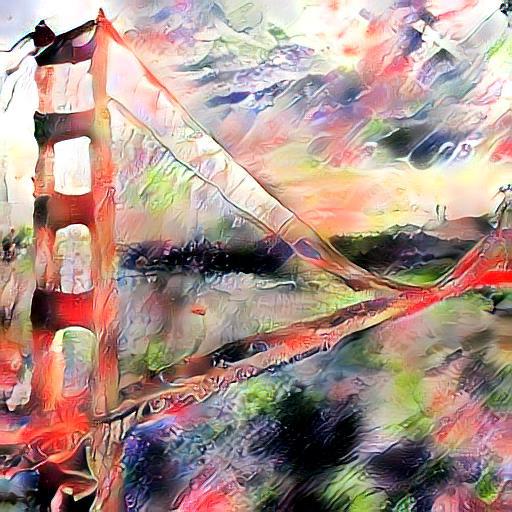}&
		\includegraphics[width=0.095\linewidth]{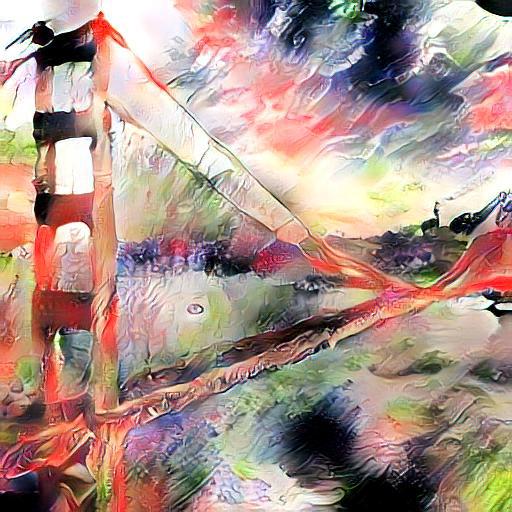}
		\\
		\footnotesize  Inputs & \multicolumn{4}{|c}{\footnotesize  (c) Avatar-Net + Our DivSwapper} &
		&&
		\footnotesize  Inputs & \multicolumn{4}{|c}{\footnotesize  (d) WCT + Our DivSwapper}
		
	\end{tabular}
	\caption{ Given the same content and style images, our proposed DivSwapper can endow existing patch-based methods (e.g., (a) CNNMRF~\protect\cite{li2016combining}, (b) Style-Swap~\protect\cite{chen2016fast}, and (c) Avatar-Net~\protect\cite{sheng2018avatar}) with the explicit ability to generate diverse stylized results. Moreover, it can also be intergated into Gram-based methods (e.g., 	(d) WCT~\protect\cite{li2017universal}) to achieve diversity. Our approach is plug-and-play and shows superiority in diversity, quality, and efficiency over the state of the art. } 
	\label{fig:teaser}
\end{figure*}

\section{Introduction}
\label{intro}

Committed to automatically transforming the style of one image to another, style transfer has become a vibrant community that attracts widespread attention from both industry and academia. The seminal work of \cite{gatys2016image} first utilized the Convolutional Neural Networks (CNNs) to extract hierarchical features and transfer the style by iteratively matching the Gram matrices (i.e., feature correlations). Since then, valuable efforts have been made to improve the efficiency~\cite{johnson2016perceptual}, quality~\cite{lin2021drafting}, and generality~\cite{huang2017arbitrary}, etc. However, as another important aspect of style transfer, {\em diversity} has received relatively less attention, and there are only a few works to solve this dilemma. \cite{li2017diversified} and \cite{ulyanov2017improved} introduced the diversity loss to train the feed-forward networks to generate diverse outputs in a learning-based mechanism. Alternatively, in a learning-free manner, \cite{wang2020diversified} proposed to use Deep Feature Perturbation based on Whitening and Coloring Transform (WCT) to perturb the deep image feature maps while keeping their Gram matrices unchanged. These methods are all Gram-based; though considerable diversity can be achieved, unfortunately, they are not applicable to other types of approaches such as patch-based methods, since these methods are not underpinned by the Gram matrix assumption.

In this work, we are interested in the diversity of the patch-based stylization mechanism. As another widespread research interest of style transfer, the patch-based method is first formulated by \cite{li2016combining,li2016precomputed}. They combined Markov Random Fields (MRFs) and CNNs to extract and match the local neural patches of the content and style images. Later, \cite{chen2016fast} proposed a Style-Swap operation and an inverse network for fast patch-based stylization. Since then, many successors were further designed for higher quality~\cite{sheng2018avatar} and extended applications~\cite{champandard2016semantic}, etc.

Let us start with a fundamental problem: {\em what limits the diversity of patch-based style transfer?} Whether using iterative optimization~\cite{li2016combining} or feed-forward networks~\cite{chen2016fast}, the core of patch-based methods is to substitute the patches of the content image with the best-matched patches of the style image (which we call ``{\bf style swapping}"~\cite{chen2016fast} in this paper), where a {\em Normalized Cross-Correlation (NCC)} approach is mainly adopted to measure the similarities of two patches. However, as we all know, the NCC heavily depends on the consistency of local variations~\cite{sheng2018avatar}, and this stereotyped patch matching process restricts each content patch to be bound to its nearest style patch, thus limiting the diversity. Though it may be effective on semantic-level style transfer (e.g., portrait-to-portrait), for more general artistic styles (e.g., Fig.~\ref{fig:teaser}), there is little semantic correspondence between them and the contents. Even for human beings, it is hard to say which patches should match best. Therefore, we argue that for {\em artistic style transfer}~\cite{gatys2016image}, it would be more reasonable to relax the restricted style swapping process and allow some meaningful variations but maintain those inherent characteristics (e.g., the approximate semantic matching). It can give users more options to select the most satisfactory results according to different preferences. Moreover, for semantic-level style transfer, the diversified matching process can also help alleviate the undesirable artifacts caused by the restricted patch matching~\cite{zhang2019multimodal} (see later Sec.~\ref{abs}).

However, making such meaningful variations is a challenging task. First, neural patches are with high dimensions and hard to control. Maybe a small change would result in significant quality degradation, or a big change might not lead to a marked visual difference~\cite{sheng2018avatar}. Therefore, the difficulty is finding the neural patches critical to visual variations and controlling them gracefully. Second, the visual effects and quality of the final results are also determined by the inherent correspondence between the content and style patches. Thus, how to manipulate this complicated correspondence to obtain diverse visual effects while maintaining the original quality is another problem to be solved.

\begin{figure}[t]
	\centering
	\includegraphics[width=1\linewidth]{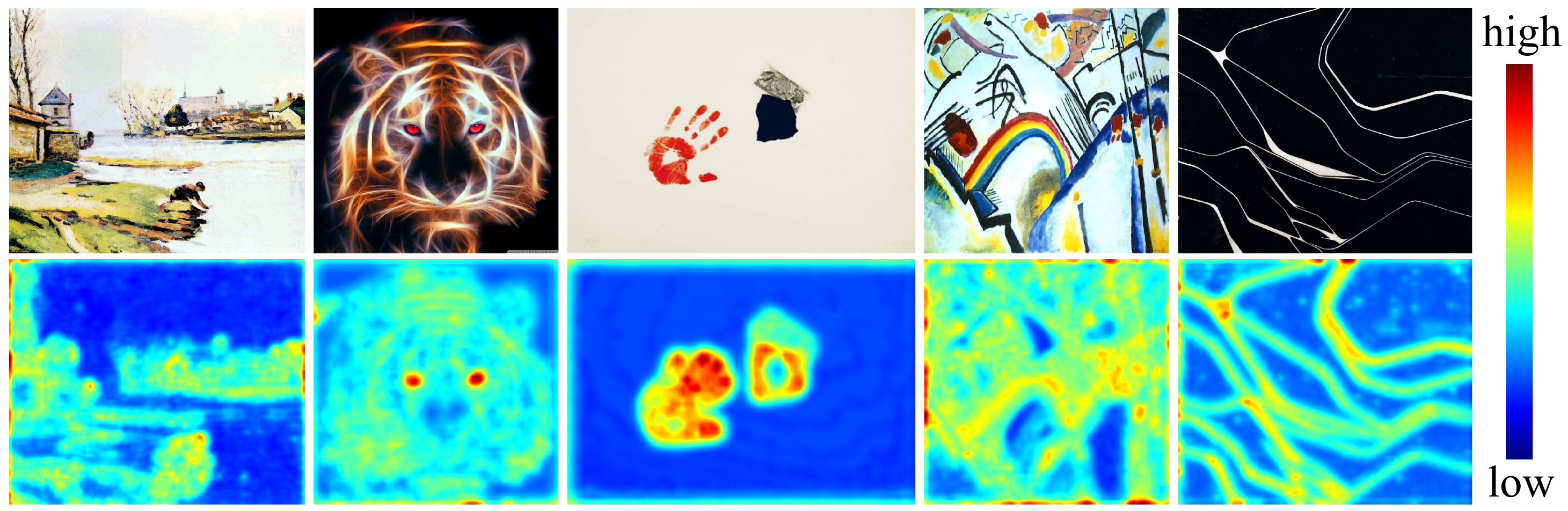}
	\caption{ Our intuition: Patches with higher activation values often contribute more to perceptually important (discriminative) information such as semantics, salient colors, and edges; thereby, they could also contribute more to diversity. Top: Some style~exemplars. Bottom: Heat maps of the activation feature maps (upsampled to the full image resolution) extracted from layer {\em Relu\_4\_1} of a pre-trained VGG19~\protect\cite{simonyan2014very}.}
	\label{fig:activation}
\end{figure}

Based on the above analyses, in this paper, we dive into the crux of patch-based style transfer and explore the universal way to diversify it. As shown in Fig.~\ref{fig:activation}, an essential intuition we will use is that the visual effects of the output images are determined by the local neural patches of the intermediate activation feature maps; since the patches with higher activation values often contribute more to perceptually important (discriminative) information~\cite{aberman2018neural,zhang2018unreasonable}, they may also contribute more to visual variations as the human eyes are often more sensitive to the changes of these parts. In other words, if we could appropriately vary these higher-activated patches, then more significant diversity can be obtained. However, directly manipulating these patches is intractable since it is hard to distinguish which patches are with higher activation values and where they should be placed so as not to degrade the quality.

To remedy it, in this work, we theoretically derive that simply shifting the L2 norm of each style patch in the style swapping process can gracefully improve diversity and vary the patches with higher activation values in an implicit and holistic way. Based on this finding, we introduce a universal and efficient module, termed {\em DivSwapper}, for diversified patch-based arbitrary style transfer. Our DivSwapper is {\em plug-and-play} and {\em learning-free}, which can be easily integrated into existing patch-based methods to help them generate diverse outputs for arbitrary styles. Besides, despite building upon the patch-based mechanism, it can also be applied to Gram-based methods to achieve higher diversity (see examples in Fig.~\ref{fig:teaser}). Theoretical analyses and extensive experiments demonstrate that our DivSwapper can achieve significant diversity while maintaining the original quality and some inherent characteristics (e.g., the approximate semantic matching) of the baseline methods. Furthermore, compared with other state-of-the-art (SOTA) diversified algorithms, it shows notable superiority in {\em diversity}, {\em quality}, and {\em efficiency}.

Overall, the main contributions of our work are threefold:
\begin{itemize}
	\setlength{\itemsep}{2pt}
	\setlength{\parsep}{0pt}
	\setlength{\partopsep}{0pt}
	\setlength{\parskip}{0pt}
	\item We explore the challenging problem of diversified patch-based style transfer and dive into its crux to achieve diversity. A universal and efficient module called {\em DivSwapper} is proposed to address the challenges and provide graceful control between diversity and quality.
	
	\item Our DivSwapper is {\em plug-and-play} and {\em learning-free}, which can be easily integrated into existing patch-based and Gram-based methods with little extra computation and time overhead.
	
	\item We analyze and demonstrate the effectiveness and superiority of our method against SOTA diversified algorithms in terms of diversity, quality, and efficiency.
\end{itemize}

\renewcommand\arraystretch{0.5}
\begin{figure*}[t]
	\centering
	\setlength{\tabcolsep}{0.05cm}
	\begin{tabular}{ccp{0.05cm}|p{0.05cm}cccp{0.05cm}|p{0.05cm}cccc}
		\includegraphics[width=0.1\linewidth]{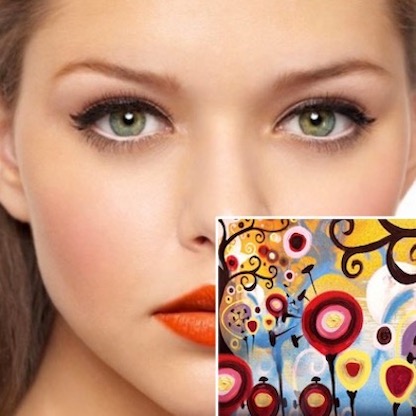}&
		\includegraphics[width=0.1\linewidth]{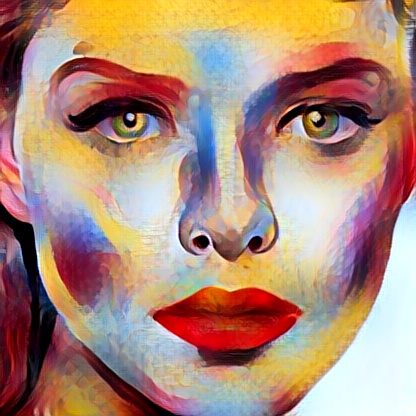}&&&
		\includegraphics[width=0.1\linewidth]{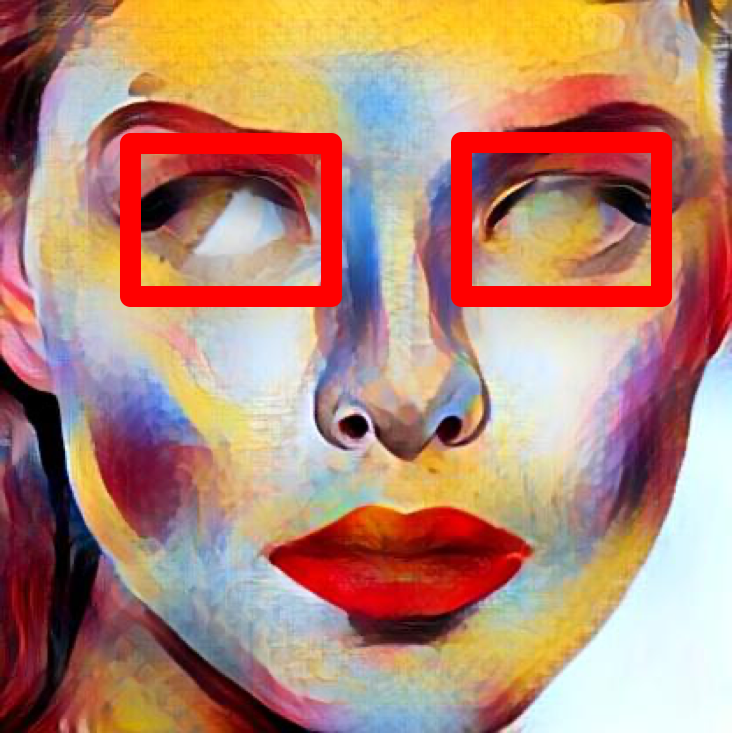}&
		\includegraphics[width=0.1\linewidth]{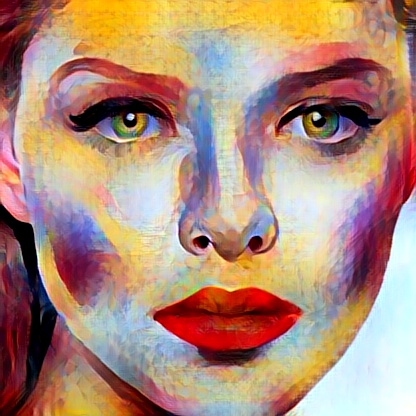}&
		\includegraphics[width=0.1\linewidth]{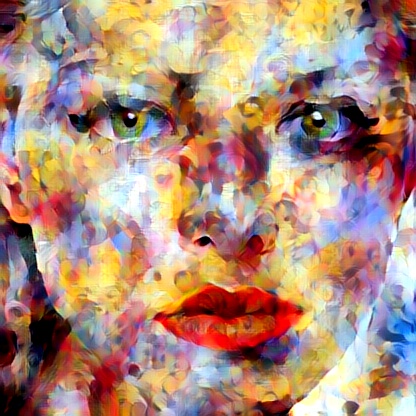}&&&
		\includegraphics[width=0.1\linewidth]{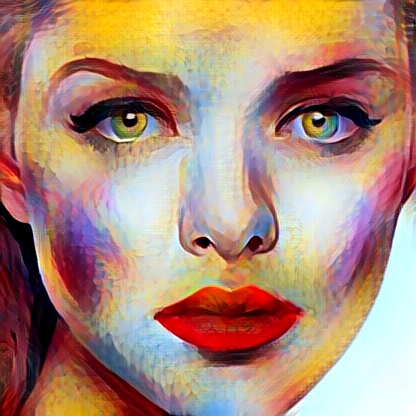} &
		\includegraphics[width=0.1\linewidth]{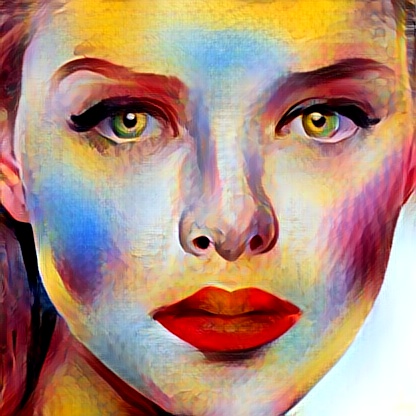} &
		\includegraphics[width=0.1\linewidth]{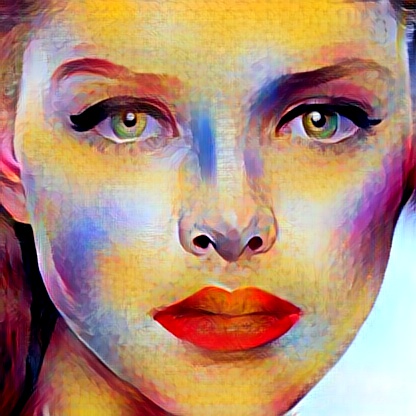} &
		\includegraphics[width=0.1\linewidth]{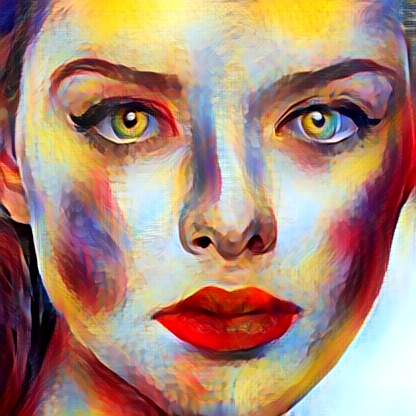}
		
		\\
		
		\footnotesize (a) Inputs &\footnotesize (b) Avatar-Net && & \footnotesize (c) + Small & \footnotesize (d) + Big & \footnotesize (e) Random &&& \multicolumn{4}{c}{  \footnotesize (f) + Our DivSwapper}
		
	\end{tabular}
	\caption{ Challenges of diversified patch-based style transfer. We adopt Avatar-Net~\protect\cite{sheng2018avatar} as the baseline method. } 
	\label{fig:challenges}
\end{figure*}

\section{Related Work}
\label{related}
The seminal work of \cite{gatys2016image} has ushered in an era of Neural Style Transfer (NST)~\cite{jing2019neural}, where the CNNs are used to decouple and recombine the styles and contents of arbitrary images. After the recent rapid development, various methods have been proposed, among which the Gram-based and patch-based are the most representative.

{\bf Gram-based methods.} The method proposed by \cite{gatys2016image} is Gram-based, using so-called Gram matrices of the feature maps extracted from CNNs to represent the styles of images, and could achieve visually stunning results. Since then, numerous Gram-based approaches were proposed to improve the performance in many aspects, including efficiency~\cite{johnson2016perceptual,ulyanov2016texture}, quality~\cite{li2017demystifying,lu2019closed,wang2020glstylenet,wang2021evaluate,lin2021drafting,chandran2021adaptive,cheng2021style,an2021artflow,chen2021dualast}, and generality~\cite{chen2017stylebank,huang2017arbitrary,li2017universal,li2019learning,jing2020dynamic}, etc.

{\bf Patch-based methods.} Patch-based style transfer is another important research line. \cite{li2016combining,li2016precomputed} first combined MRFs and CNNs for arbitrary style transfer. It extracts local neural patches to represent the styles of images and searches for the most similar patches from the style image to satisfy the local structure prior of the content image. Later, \cite{chen2016fast} proposed to swap the content activation patch with the best-matched style activation patch using a Style-Swap operation, and then used an inverse network for fast patch-based stylization. Based on them, many successors were further designed for better performance~\cite{sheng2018avatar,gu2018arbitrary,park2019arbitrary,kolkin2019style,yao2019attention,zhang2019multimodal,deng2020arbitrary,liu2021adaattn,chen2021artistic} and extended applications~\cite{champandard2016semantic,liao2017visual,wang2022texture}.

{\bf Diversified methods.} Our method is closely related to the existing diversified methods. \cite{li2017diversified} and \cite{ulyanov2017improved} introduced the diversity loss to train the feed-forward networks to generate diverse outputs by mutually comparing and maximizing the variations between the generated results in mini-batches. However, these methods are learning-based and have restricted generalization, limited diversity, and poor scalability~\cite{wang2020diversified}. To combat these limitations, \cite{wang2020diversified} proposed a learning-free method called Deep Feature Perturbation to empower the WCT-based methods to generate diverse results. This method is universal for arbitrary styles, but unfortunately, it relies on WCT and does not apply to other types of methods. 

{\bf Discussions.} While there have been some efforts for the diversified style transfer, they are all Gram-based and are not applicable to other types of approaches such as patch-based methods. As another widespread research interest, the diversity of patch-based style transfer remains challenging. {\em Our work, as far as we know, takes the first step in this direction}. The proposed approach is learning-free and universal for arbitrary styles, and can be easily embedded into existing patch-based methods to empower them to generate diverse results. Moreover, it can also be applied to Gram-based methods to achieve higher diversity. Compared with the state of the art, our approach can achieve higher diversity, quality, and efficiency, which will be validated in later Sec.~\ref{exp}.

\section{Proposed Approach}
\label{approach}
Before introducing our approach, let us first reiterate {\em why implementing diversity in patch-based methods is challenging?} 

First, neural patches are with high dimensions and hard to control. On the one hand, maybe a small change would easily result in significant quality degradation, e.g., Fig.~\ref{fig:challenges} (c). As can be observed in the red box areas, the result exhibits a severe quality problem that the portrait's eyes disappear, even if we only change 50 of the total 2500 neural patches. On the other hand, it is also possible that a big change may not lead to a marked visual difference, e.g., Fig.~\ref{fig:challenges} (d). Although all the 2500 neural patches have been changed, the result is still very similar to the original one in Fig.~\ref{fig:challenges} (b). Therefore, the difficulty is finding the neural patches {\em critical} to visual variations and controlling them {\em gracefully}. 

Second, the inherent correspondence between the content and style patches ensures the visual correctness and rationality of the final results, as well as the semantic correspondence. If we simply ignore this local correspondence (e.g., randomly matching a style patch for each content patch), it will destroy the content prior and generate poor results, as shown in Fig.~\ref{fig:challenges}~(e). Therefore, one {\em key desideratum} of the diversified patch-based style transfer is to generate meaningful variations while maintaining the original quality and some inherent characteristics (e.g., the approximate semantic matching). 

Aiming at the challenges above and based on the intuition introduced in Sec.~\ref{intro}, we propose a simple yet effective diversified style swapping module, termed {\em DivSwapper}, for diversified patch-based arbitrary style transfer. The proposed module is plug-and-play and learning-free, which can be easily integrated into existing patch-based and Gram-based methods to achieve diversity. As shown in Fig.~\ref{fig:challenges}~(f), the synthesized diverse results are all reasonable, with meaningful variations while maintaining the original quality and some inherent characteristics (e.g., the approximate semantic matching). In the following section, we will first describe the workflow of our proposed DivSwapper, and then introduce the key finding and design in DivSwapper to achieve diversity, i.e., the {\em Shifted Style Normalization (SSN)}. In the light of this, we theoretically derive its effectiveness in generating diverse reasonable solutions and helping vary more significant neural patches with higher activation values.

\begin{figure}[t]
	\centering
	\includegraphics[width=1\linewidth]{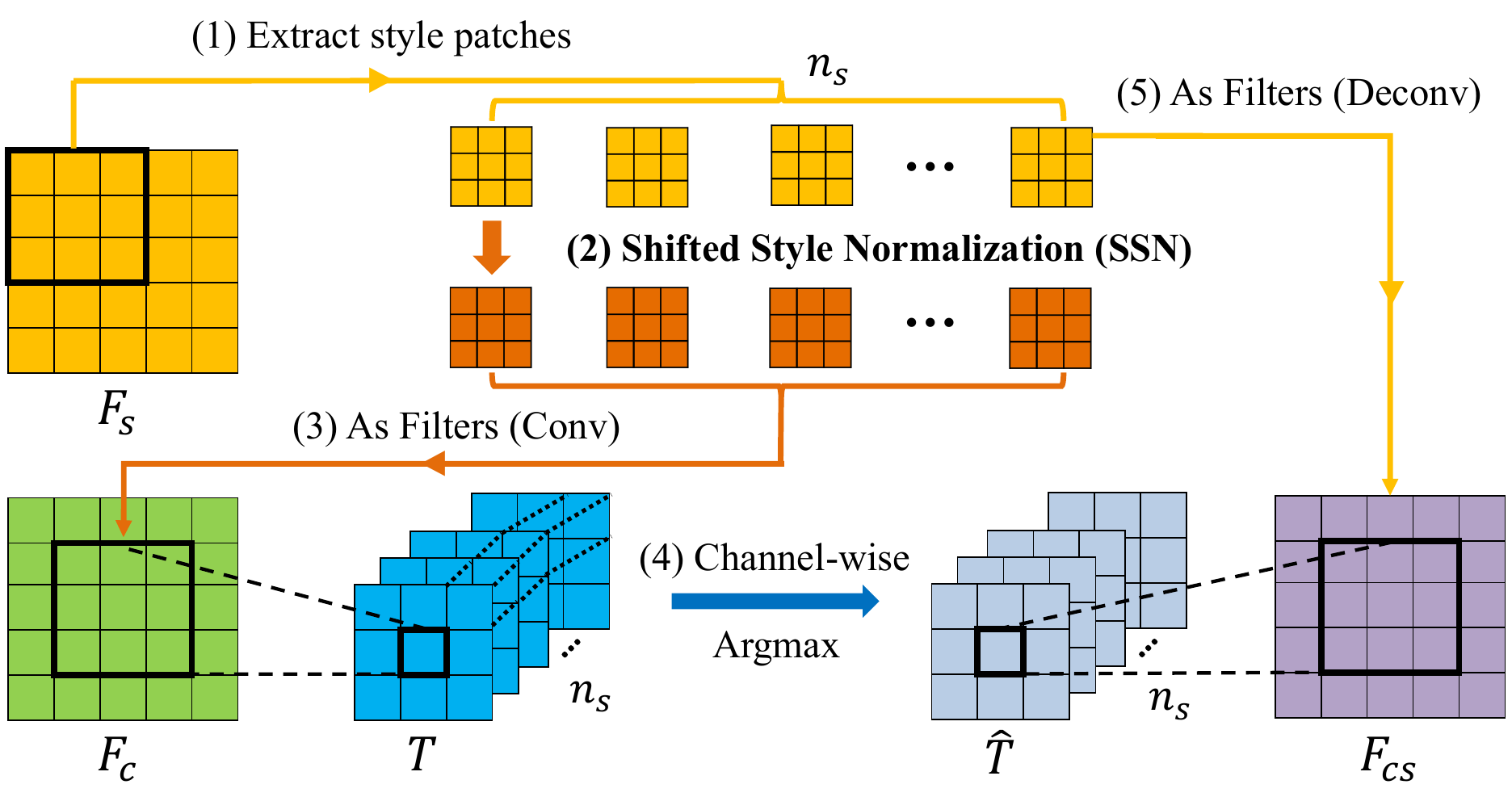}
	\caption{ The workflow of our proposed DivSwapper.
	}
	\label{fig:over}
\end{figure}

\subsection{Workflow of DivSwapper}
\label{basic}
Given a content image $I_c$ and style image $I_s$ pair, suppose $F_c=CNN(I_c)$ and $F_s=CNN(I_s)$ are content and style activation feature maps extracted from a certain layer (e.g., {\em Relu\_4\_1}) of a pre-trained CNN (e.g., VGG~\cite{simonyan2014very}). As shown in Fig.~\ref{fig:over} (step (1-5)), our DivSwapper aims to search for the diverse yet plausible style patches in $F_s$ for each content patch in $F_c$, and then substitute the latter with the former. The detailed workflow is:

\begin{enumerate}
	\setlength{\itemsep}{2pt}
	\setlength{\parsep}{0pt}
	\setlength{\partopsep}{0pt}
	\setlength{\parskip}{0pt}
	\renewcommand{\labelenumi}{(\theenumi)}
	\item Extract the style patches from $F_s$, denoted as $\{\phi_j(F_s)\}_{j\in \{1,\dots,n_s\}}$, where $n_s$ is the number of patches.
	
	\item Normalize each style patch by using a {\em Shifted Style Normalization (SSN)} approach. The shifted normalized style patches are denoted as $\{\hat{\phi}_j(F_s)\}$.
	
	\item Calculate the similarities between all pairs of the style and content patches by the {\em Normalized Cross-Correlation (NCC)} measure, i.e, $\mathcal{S}_{i,j}=\langle \phi_i(F_c),\hat{\phi}_j(F_s)\rangle$ (the norm of the content patch $\phi_i(F_c)$ is removed as it is constant with respect to the $\mathop{\arg\max}$ operation in the next step (4)). This process can be efficiently implemented by using a convolutional layer with the shifted normalized style patches $\{\hat{\phi}_j(F_s)\}$ as filters and content feature map $F_c$ as input. The computed result $T$ has $n_s$ feature channels, and each spatial location is a vector of NCC between a content patch and all style patches.
	
	\item Find the nearest style patch for each content patch, i.e., $\phi_i(F_{cs})=\mathop{\arg\max}_{j\in\{1,\dots,n_s\}} \mathcal{S}_{i,j}$. It can be achieved by first finding the channel-wise argmax for each spatial location of $T$, and then replacing it with a channel-wise one-hot encoding. The result is denoted as $\hat T$.
	
	\item Reconstruct the swapped feature ${ F_{cs}}$ by a deconvolutional layer with the {\em original style patches} $\{\phi_j(F_s)\}$ as filters and $\hat T$ as input.
\end{enumerate}

{\bf Analysis:} The most novel insight behind DivSwapper is that we use a {\em SSN} approach to inject diversity into the {\em NCC-based} style swapping process, which kills three birds with one stone: i) We can reshuffle {\em all} style patches by adding random norm shifts, which ensures the scope of diversity. ii) NCC is still used for nearest patch matching, and the final swapped feature ${ F_{cs}}$ is reconstructed by the {\em original style patches}, thereby the original quality and the inherent characteristics (e.g., the approximate semantic matching) can be well maintained. iii) SSN implicitly helps vary more significant style patches with higher activation values, thus achieving more meaningful diversity (see more analyses in Sec.~\ref{method}). {\em Note that since the matching step (3) and the reconstruction step (5) actually can be implemented by two convolutional layers, our DivSwapper is very efficient.}

\subsection{Shifted Style Normalization}
\label{method}

\begin{figure}[t]
	\centering
	\includegraphics[width=0.9\linewidth]{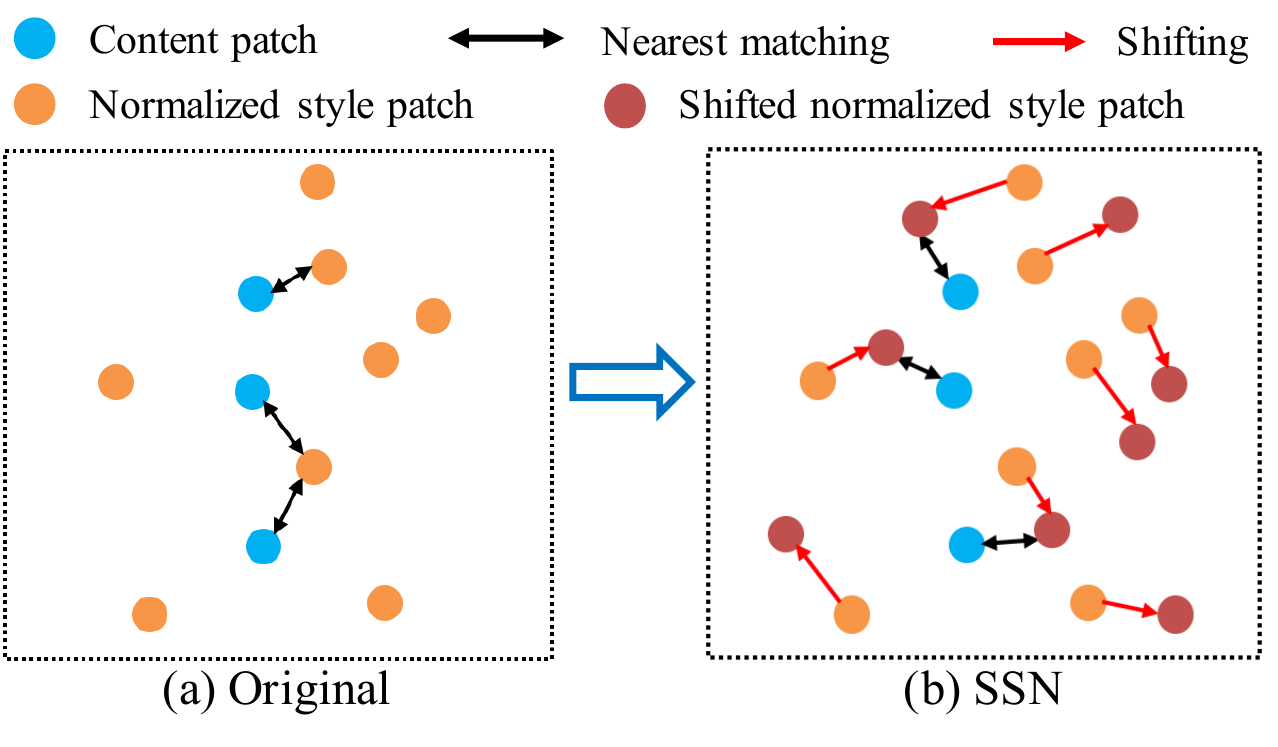}
	\caption{(a) The original style swapping process only produces one deterministic solution by matching the nearest style patch with each content patch. (b) Our SSN adds random deviations to shift the style patch normalization, thus achieving diversity.
	}
	\label{fig:ops}
\end{figure}

The stereotyped style swapping process aims to search for the nearest style patch for each content patch, which only produces one deterministic solution, as illustrated in Fig.~\ref{fig:ops}~(a). To obtain different solutions, an intuitive way is to match other plausible style patches instead of the nearest ones, which can be achieved by adjusting the distances between the content and style patches. However, as analyzed in Sec.~\ref{intro}, the key to obtaining more meaningful diversity is to gracefully control and vary those significant patches with higher activation values. Therefore, we propose the {\em Shifted Style Normalization (SSN)} to explicitly alter the distances between the content and style patches while implicitly restricting the swapping process to vary more significant style patches with higher activation values.

Simply yet non-trivially, as illustrated in Fig.~\ref{fig:ops} (b), our SSN adds a random {\em positive} deviation $\sigma$ to shift the L2 norm of {\bf each} style patch, like follows: 
\begin{equation}
\{\hat{\phi}_j(F_s)=\frac{\phi_j(F_s)}{\parallel \phi_j(F_s) \parallel+\sigma}\}_{j\in \{1,\dots,n_s\}}.
\end{equation}

Now, we theoretically derive the power of this {\em ``magical''} random deviation $\sigma$ to generate diverse solutions and help gracefully vary more significant style patches with higher activation values. For simplicity, we only take one content and two style activation patches to illustrate, which are denoted as $\mathcal{P}^c$, $\mathcal{P}^s_1$, and $\mathcal{P}^s_2$, respectively. Note that the values in these vectors are {\em non-negative} because they are often extracted from the ReLU activation layers (e.g., {\em Relu\_4\_1}) of VGG model. Specifically, we first suppose that they satisfy the following original NCC matching relationship:
\begin{equation}
\frac{\langle \mathcal{P}^c,\mathcal{P}^s_1\rangle}{\parallel \mathcal{P}^c\parallel \parallel \mathcal{P}^s_1\parallel} = \cos \theta_1 > \frac{\langle \mathcal{P}^c,\mathcal{P}^s_2\rangle}{\parallel \mathcal{P}^c\parallel \parallel \mathcal{P}^s_2\parallel} = \cos \theta_2 > 0,
\label{eq3}
\end{equation}
which means $\mathcal{P}^s_1$ matches $\mathcal{P}^c$ better than $\mathcal{P}^s_2$, where $\theta_1$ is the angle between vector $\mathcal{P}^c$ and $\mathcal{P}^s_1$, $\theta_2$ is the angle between vector $\mathcal{P}^c$ and $\mathcal{P}^s_2$. We want to change their matching relationship by randomly shifting the L2 norms of the style patches, i.e.,
\begin{equation}
\frac{\langle \mathcal{P}^c,\mathcal{P}^s_1\rangle}{\parallel \mathcal{P}^c\parallel (\parallel \mathcal{P}^s_1\parallel +\sigma_1)} < \frac{\langle \mathcal{P}^c,\mathcal{P}^s_2\rangle}{\parallel \mathcal{P}^c\parallel (\parallel \mathcal{P}^s_2\parallel +\sigma_2)}.
\label{eq4}
\end{equation}
Thus, we can deduce:
\begin{equation}
\begin{aligned}
\langle \mathcal{P}^c,\mathcal{P}^s_2\rangle  \sigma_1  -\langle \mathcal{P}^c,\mathcal{P}^s_1\rangle \sigma_2  > \\
\langle \mathcal{P}^c,\mathcal{P}^s_1\rangle  \parallel \mathcal{P}^s_2\parallel   -  \langle \mathcal{P}^c,\mathcal{P}^s_2\rangle \parallel \mathcal{P}^s_1\parallel.
\end{aligned}
\end{equation}
Since {\small $\langle \mathcal{P}^c,\mathcal{P}^s_1\rangle \parallel \mathcal{P}^s_2\parallel - \langle \mathcal{P}^c,\mathcal{P}^s_2\rangle \parallel \mathcal{P}^s_1\parallel >0$} (Eq.~(\ref{eq3})), we can get the following solution:
$$ \langle \mathcal{P}^c,\mathcal{P}^s_2\rangle  \sigma_1  -\langle \mathcal{P}^c,\mathcal{P}^s_1\rangle \sigma_2 > 0 \Rightarrow  \langle \mathcal{P}^c,\mathcal{P}^s_2\rangle \sigma_1 > \langle \mathcal{P}^c,\mathcal{P}^s_1\rangle \sigma_2.$$
As $\sigma_1$ and $\sigma_2$ are positive and i.i.d. (independent and identically distributed), it turns out that {\em holistically} our SSN {\em tends} to replace $\mathcal{P}^s_1$ with a suitable $\mathcal{P}^s_2$ which satisfies $\langle \mathcal{P}^c,\mathcal{P}^s_2\rangle > \langle \mathcal{P}^c,\mathcal{P}^s_1\rangle$. Since $\langle \mathcal{P}^c,\mathcal{P}^s_2\rangle = \parallel \mathcal{P}^c \parallel \parallel \mathcal{P}^s_2 \parallel \cos \theta_2$, and $\langle \mathcal{P}^c,\mathcal{P}^s_1\rangle = \parallel\mathcal{P}^c \parallel \parallel \mathcal{P}^s_1 \parallel \cos \theta_1$, we can deduce as follows:
\begin{equation}
\parallel \mathcal{P}^s_2 \parallel \cos \theta_2 > \parallel \mathcal{P}^s_1 \parallel \cos \theta_1 \Rightarrow  \frac{\parallel \mathcal{P}^s_2 \parallel}{\parallel \mathcal{P}^s_1 \parallel} > \frac{\cos \theta_1}{\cos \theta_2}.
\end{equation}
As $\cos \theta_1 > \cos \theta_2$ (Eq.~(\ref{eq3})), we can obtain $\parallel\mathcal{P}^s_2\parallel > \parallel \mathcal{P}^s_1 \parallel$, which means the varied $\mathcal{P}^s_2$ often has higher activation values than original $\mathcal{P}^s_1$. That is to say, our SSN could help vary more significant style patches with higher activation values in an implicit and holistic way. Besides, since it is still implicitly constrained by the original NCC (Eq.~(\ref{eq3})) and the variations are gracefully controlled by the sampling range of $\sigma$, the overall quality and approximate semantic matching can be well preserved, as will be demonstrated in later Sec.~\ref{abs}.

\renewcommand\arraystretch{0.5}
\begin{figure}[t!]
	\centering
	\setlength{\tabcolsep}{0.015cm}
	\begin{tabular}{cp{0.02cm}|p{0.02cm}ccccc}
		\includegraphics[width=0.16\linewidth]{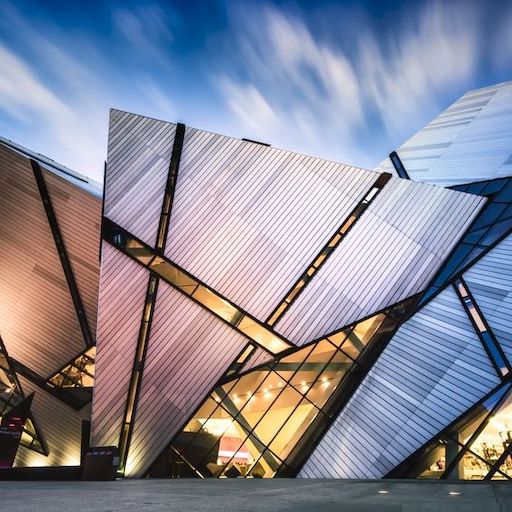}&&&
		\includegraphics[width=0.16\linewidth]{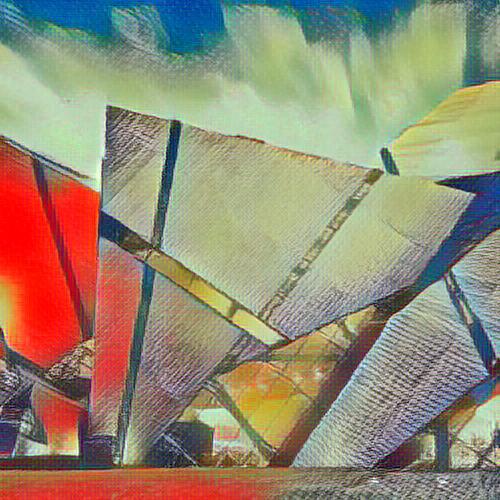}&
		\includegraphics[width=0.16\linewidth]{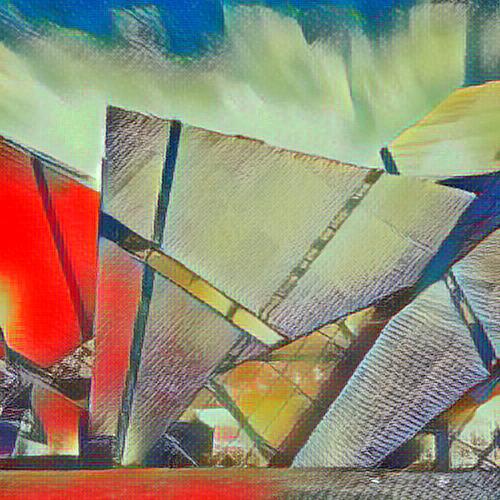}&
		\includegraphics[width=0.16\linewidth]{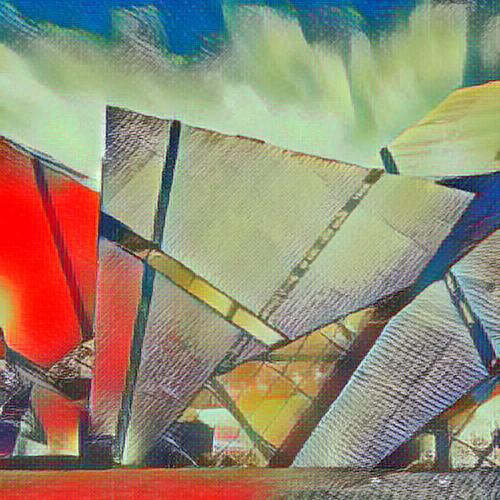}&
		\includegraphics[width=0.16\linewidth]{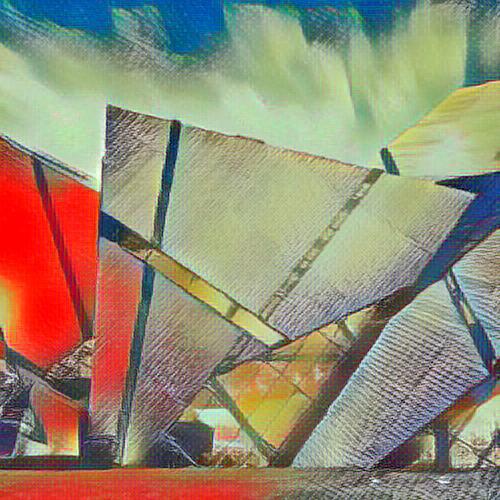}&
		\includegraphics[width=0.16\linewidth]{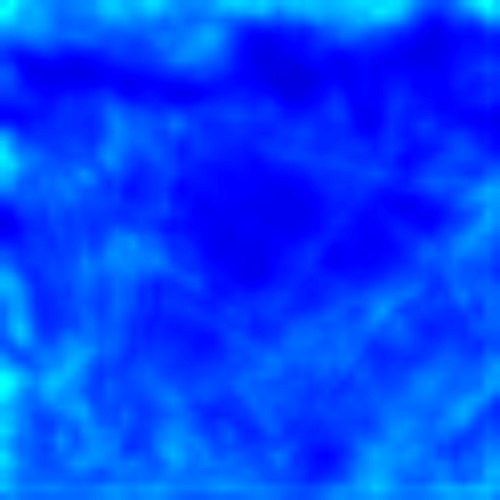}
		\\
		\scriptsize Content  &&&  \multicolumn{5}{c}{ \scriptsize (a) MTS}
		\\
		
		\includegraphics[width=0.12\linewidth]{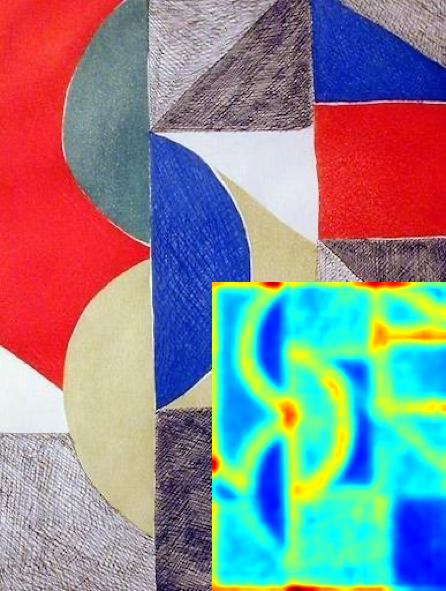}&&&
		\includegraphics[width=0.16\linewidth]{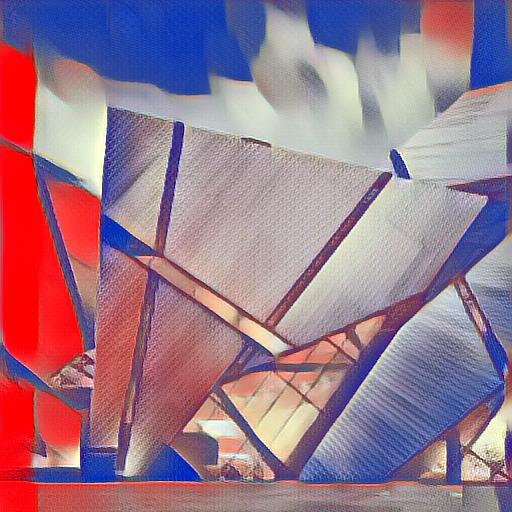}&
		\includegraphics[width=0.16\linewidth]{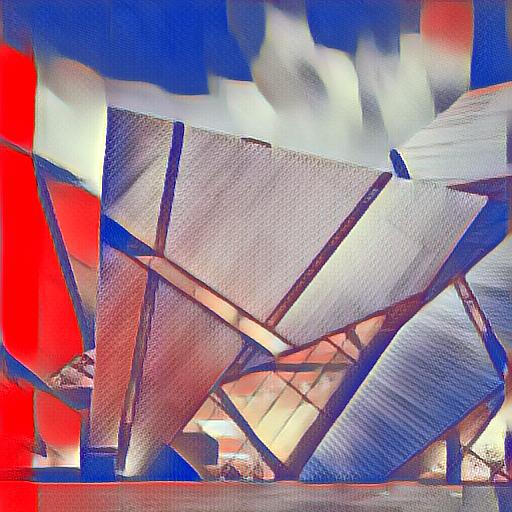}&
		\includegraphics[width=0.16\linewidth]{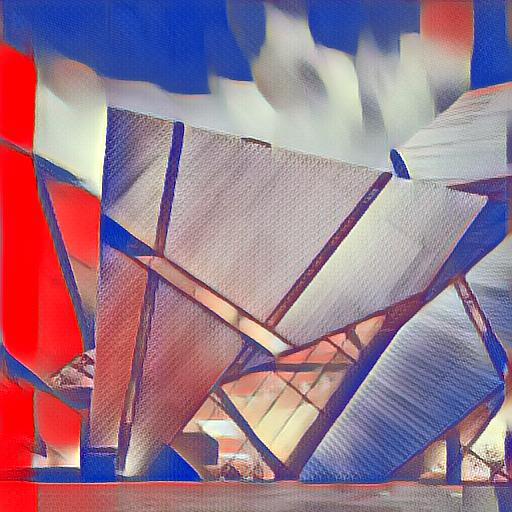}&
		\includegraphics[width=0.16\linewidth]{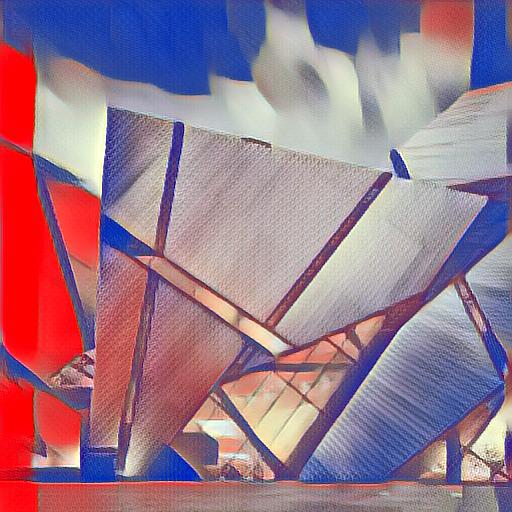}&
		\includegraphics[width=0.16\linewidth]{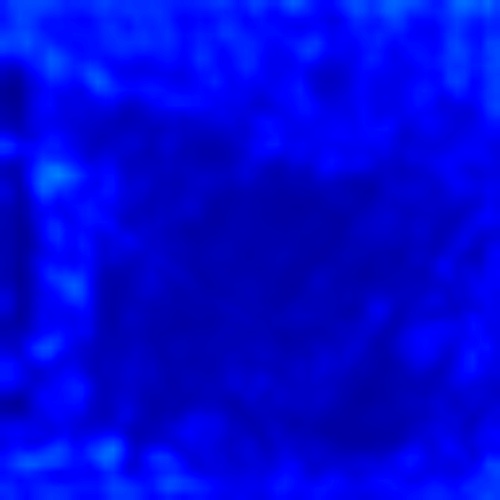}
		\\
		\scriptsize Style  &&&  \multicolumn{5}{c}{ \scriptsize (b) ITN}
		\\
		
		\includegraphics[width=0.16\linewidth]{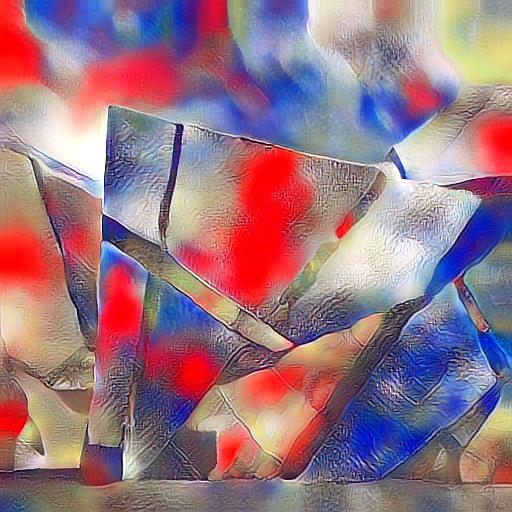}&&&
		\includegraphics[width=0.16\linewidth]{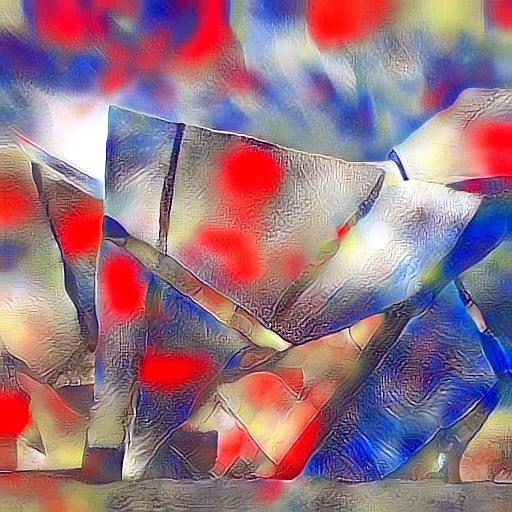}&
		\includegraphics[width=0.16\linewidth]{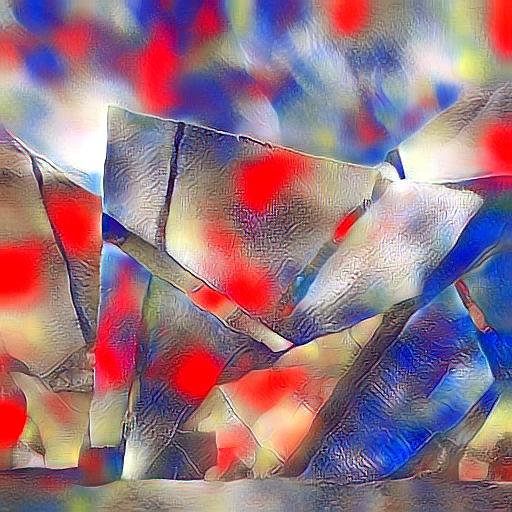}&
		\includegraphics[width=0.16\linewidth]{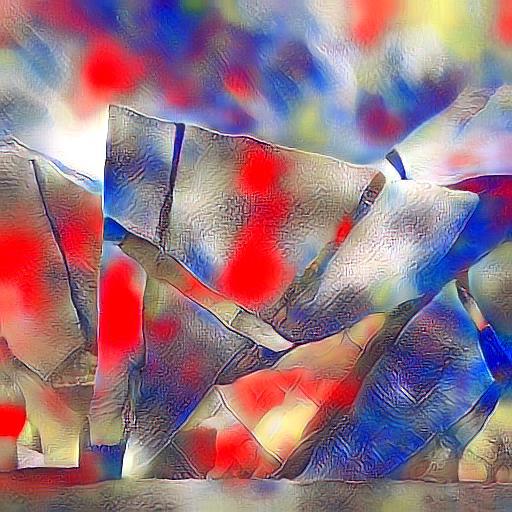}&
		\includegraphics[width=0.16\linewidth]{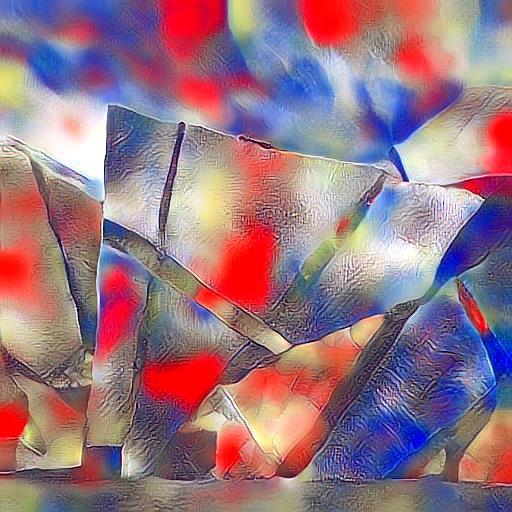}&
		\includegraphics[width=0.16\linewidth]{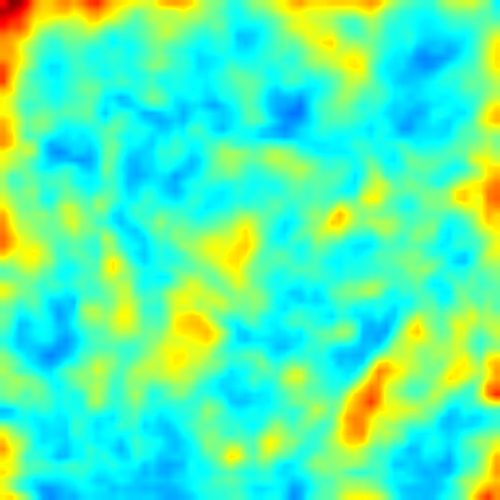}
		\\
		\scriptsize WCT  &&&  \multicolumn{5}{c}{ \scriptsize (c) WCT + DFP}
		\\
		
		\includegraphics[width=0.16\linewidth]{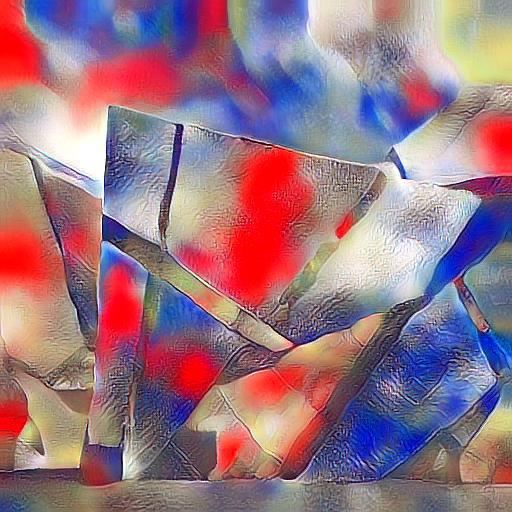}&&&
		\includegraphics[width=0.16\linewidth]{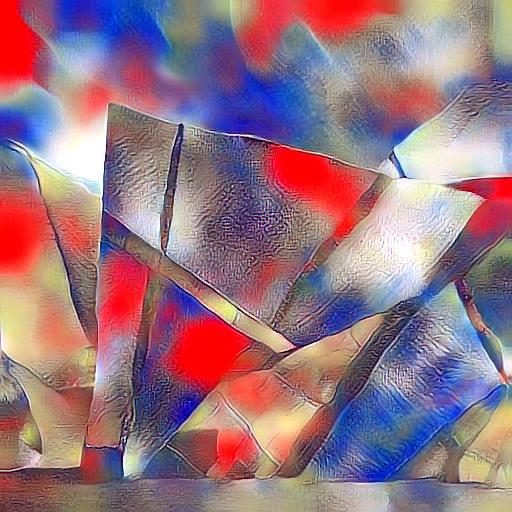}&
		\includegraphics[width=0.16\linewidth]{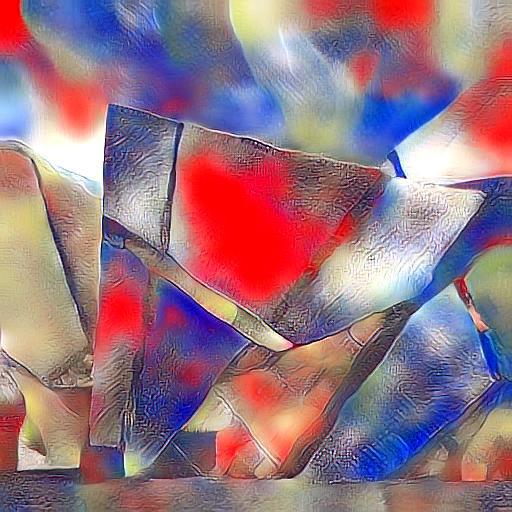}&
		\includegraphics[width=0.16\linewidth]{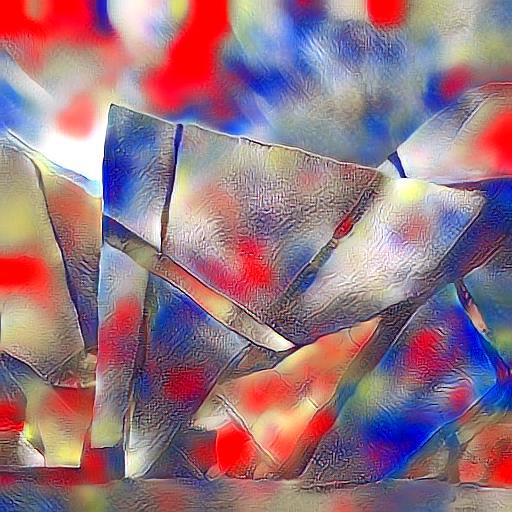}&
		\includegraphics[width=0.16\linewidth]{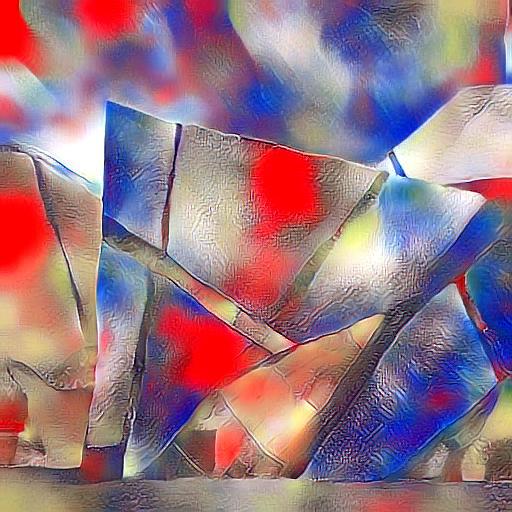}&
		\includegraphics[width=0.16\linewidth]{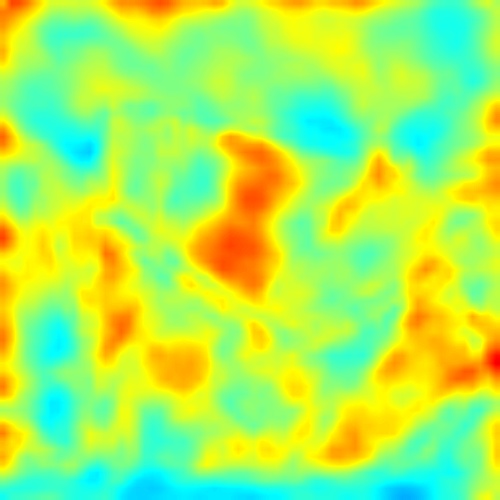}
		\\
		\scriptsize WCT  &&&  \multicolumn{5}{c}{ \scriptsize (d) WCT + \bf  Our DivSwapper}
		\\
		
		\includegraphics[width=0.16\linewidth]{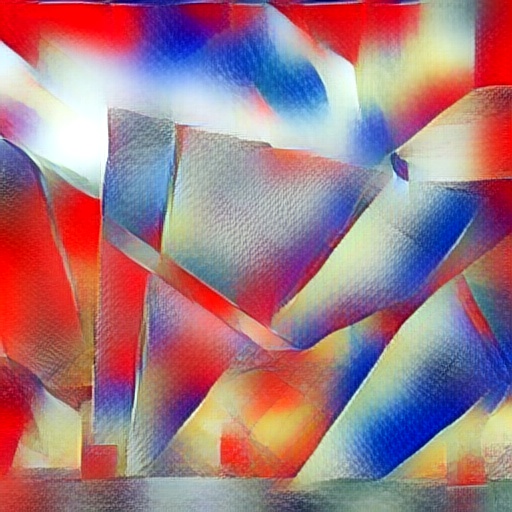}&&&
		\includegraphics[width=0.16\linewidth]{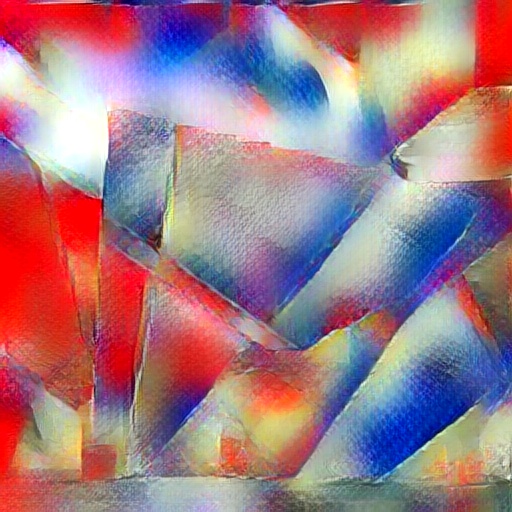}&
		\includegraphics[width=0.16\linewidth]{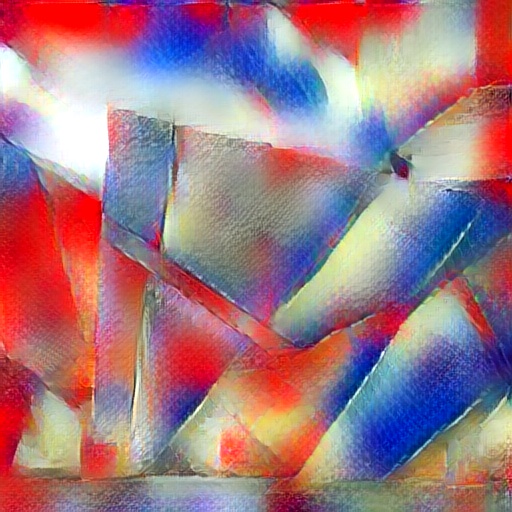}&
		\includegraphics[width=0.16\linewidth]{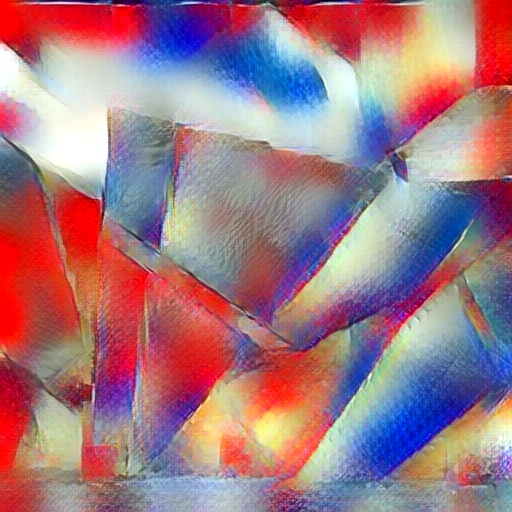}&
		\includegraphics[width=0.16\linewidth]{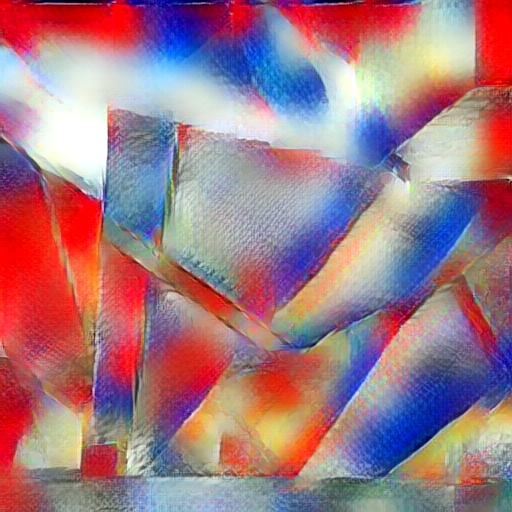}&
		\includegraphics[width=0.16\linewidth]{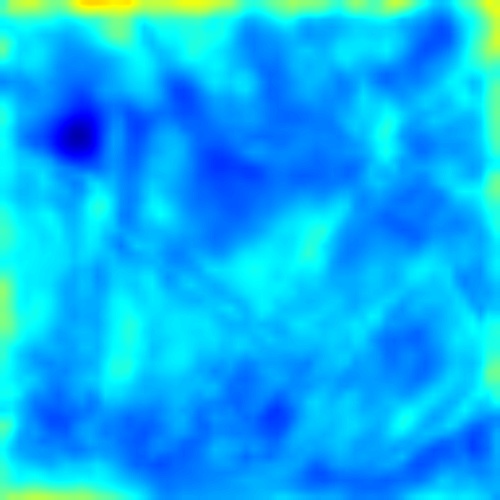}
		\\
		\scriptsize Avatar-Net  &&&  \multicolumn{5}{c}{ \scriptsize (e) Avatar-Net + DFP}
		\\
		
		\includegraphics[width=0.16\linewidth]{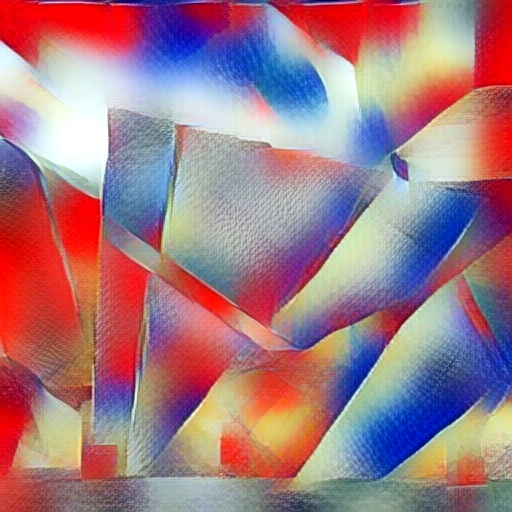}&&&
		\includegraphics[width=0.16\linewidth]{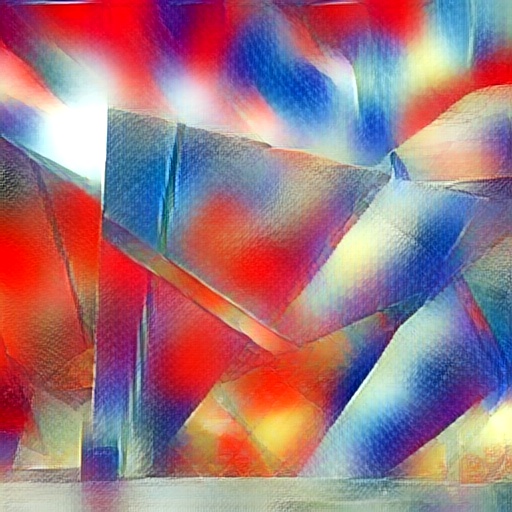}&
		\includegraphics[width=0.16\linewidth]{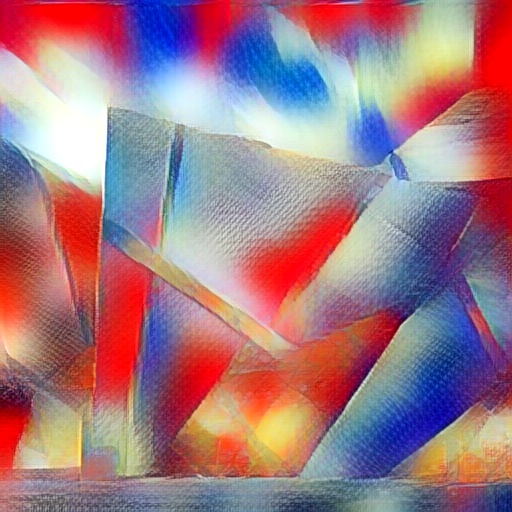}&
		\includegraphics[width=0.16\linewidth]{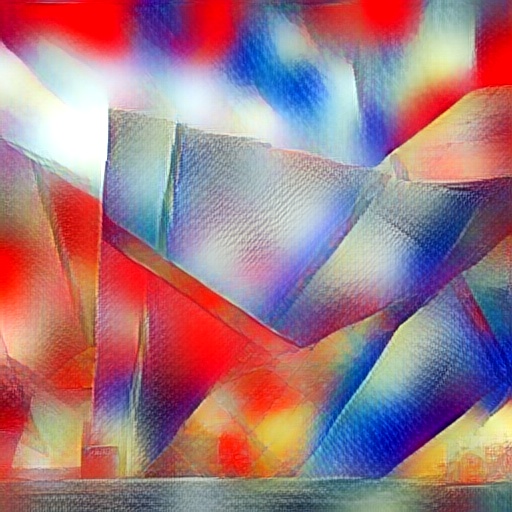}&
		\includegraphics[width=0.16\linewidth]{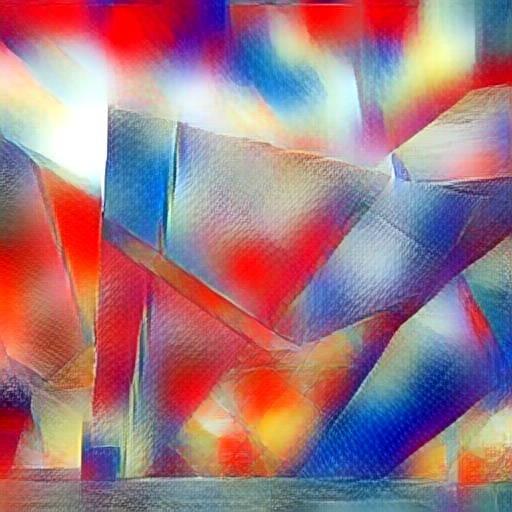}&
		\includegraphics[width=0.16\linewidth]{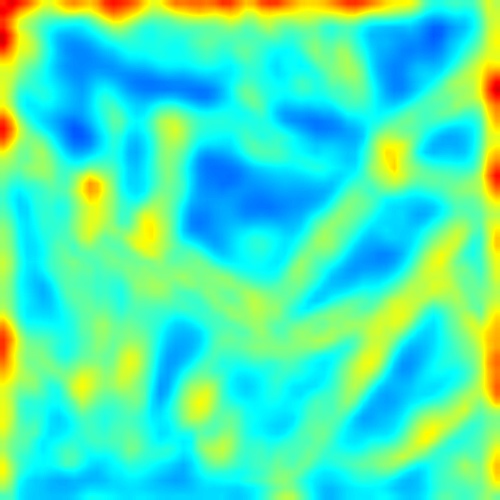}
		\\
		\scriptsize Avatar-Net  &&&  \multicolumn{5}{c}{ \scriptsize (f) Avatar-Net + \bf  Our DivSwapper}
		
	\end{tabular}
	\caption{Qualitative comparisons. From top to bottom, the first column shows the input content and style images and the original outputs of baselines; the middle four columns show the diverse outputs of MTS, ITN, baselines + DFP, and baselines + our DivSwapper [{\em best viewed in color changes and zoomed-in}]. We also visualize their average activation feature differences (averaged on $C_{20}^2 = 190$ pairs of diverse results) via heat maps in the last column.} 
	\label{fig:quality}
\end{figure}

\section{Experimental Results}
\label{exp}
\subsection{Implementation Details} 
\label{detail}
{\bf Baselines.} We integrate our DivSwapper into two types of patch-based methods based on (1) iteration optimization (CNNMRF~\cite{li2016combining}) and (2) feed-forward networks (Style-Swap~\cite{chen2016fast} and Avatar-Net~\cite{sheng2018avatar}). Besides, we also integrate it into a typical Gram-based method, i.e., WCT~\cite{li2017universal}. We keep the default settings of these baselines and fine-tune the sampling range of our $\sigma$ (sampled from a {\em uniform} distribution) to make our quality similar to the baselines, i.e., $(0, 10^3]$ for CNNMRF, $(0, 10^5]$ for Style-Swap, $(0, 5\times10^3]$ for Avatar-Net, and $(0, 5\times10^3]$ for WCT. We will discuss these settings in later Sec.~\ref{abs}. For more implementation details, please refer to the {\em supplementary material (SM)}.

{\bf Metrics.} To evaluate the diversity, we collect 36 content-style pairs released by~\cite{wang2020diversified}. For each pair, we randomly produce 20 outputs, so there are a total of $36 \times C_{20}^2=6840$ pairs of outputs generated by each method. Like \cite{wang2020diversified}, we adopt the average pixel distance $D_{pixel}$ and LPIPS ({\em Learned Perceptual Image Patch Similarity}) distance $D_{LPIPS}$~\cite{zhang2018unreasonable} to measure the diversity in pixel space and deep feature space, respectively.

\subsection{Comparisons with Prior Arts}
\label{comp}

We compare our DivSwapper with three SOTA diversified methods, i.e., Multi-Texture-Synthesis (MTS)~\cite{li2017diversified}, Improved-Texture-Networks (ITN)~\cite{ulyanov2017improved}, and Deep-Feature-Perturbation (DFP)~\cite{wang2020diversified}. Since these methods are all Gram-based, we integrate our DivSwapper into the Gram-based baseline WCT~\cite{li2017universal} and the Gram-and-patch-based baseline Avatar-Net~\cite{sheng2018avatar} for a fair comparison. 

\begin{figure}[t!]
	\centering
	\setlength{\tabcolsep}{0.015cm}
	\begin{tabular}{ccp{0.02cm}|p{0.02cm}cccc}
		
		\includegraphics[width=0.16\linewidth]{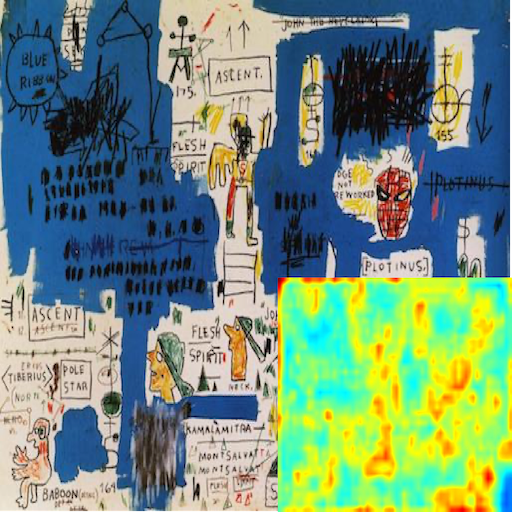}&
		\includegraphics[width=0.16\linewidth]{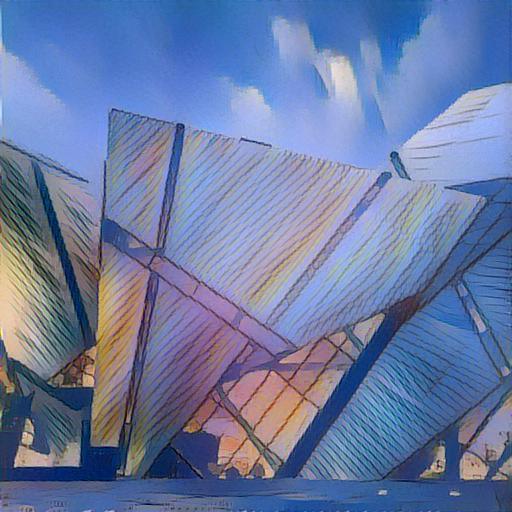}&&&
		\includegraphics[width=0.16\linewidth]{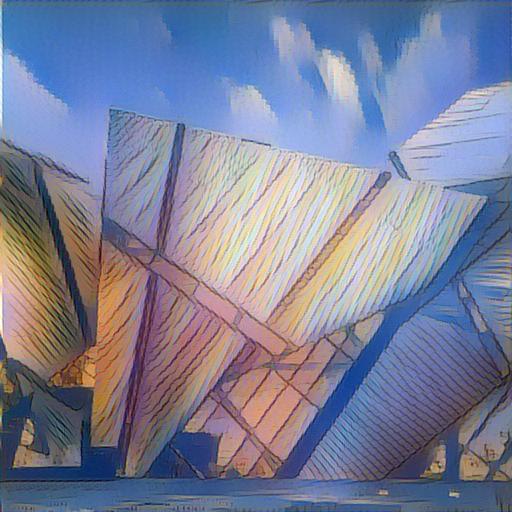}&
		\includegraphics[width=0.16\linewidth]{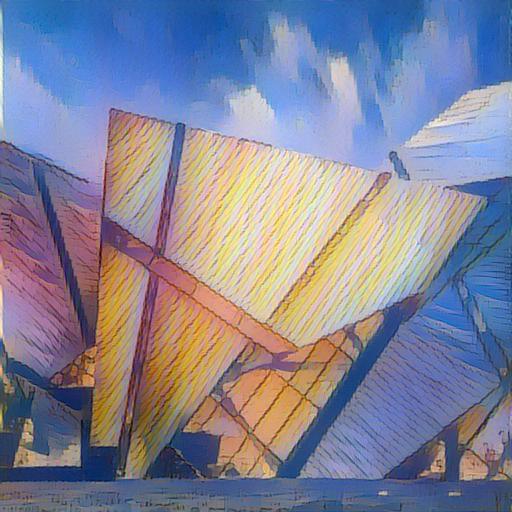}&
		\includegraphics[width=0.16\linewidth]{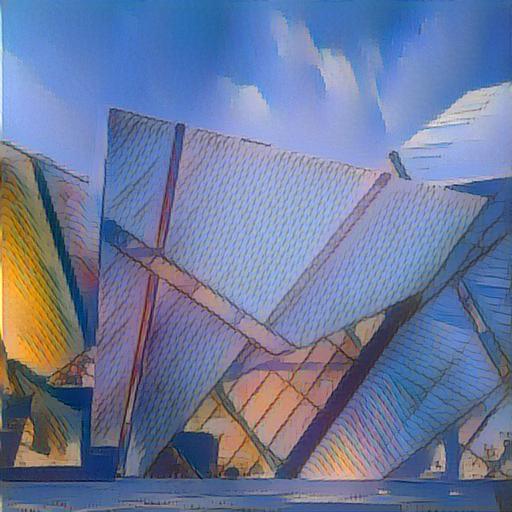}&
		\includegraphics[width=0.16\linewidth]{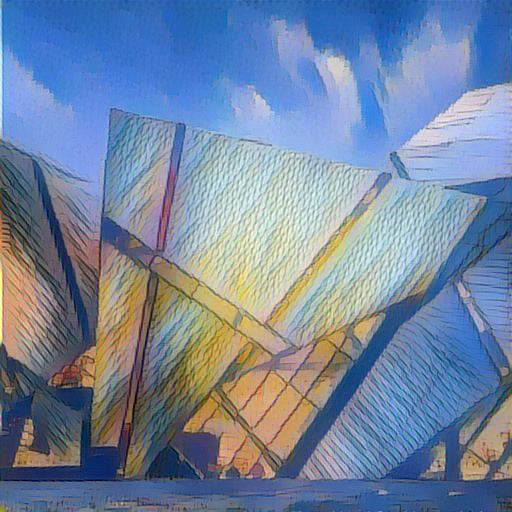}
		\\
		\scriptsize Style &\scriptsize Style-Swap &&&  \multicolumn{4}{c}{ \scriptsize Style-Swap + \bf  Our DivSwapper}
		\\
		
		\includegraphics[width=0.121\linewidth]{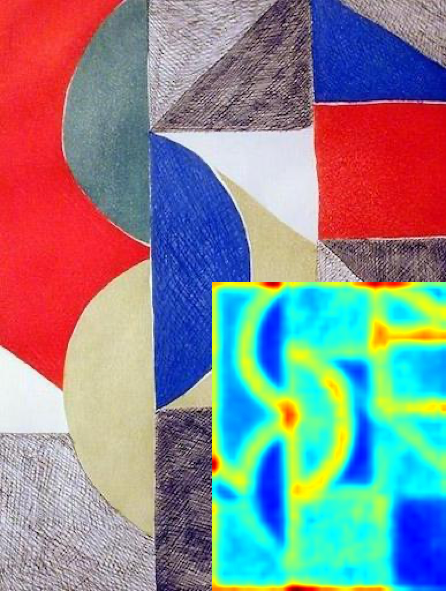}&
		\includegraphics[width=0.16\linewidth]{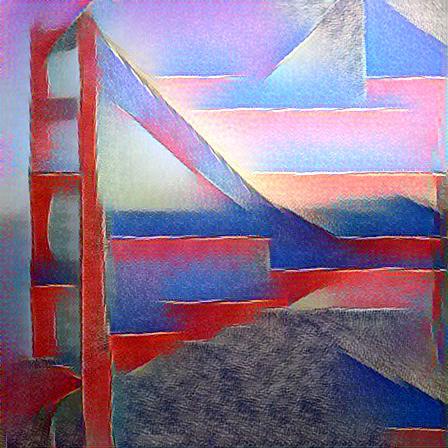}&&&
		\includegraphics[width=0.16\linewidth]{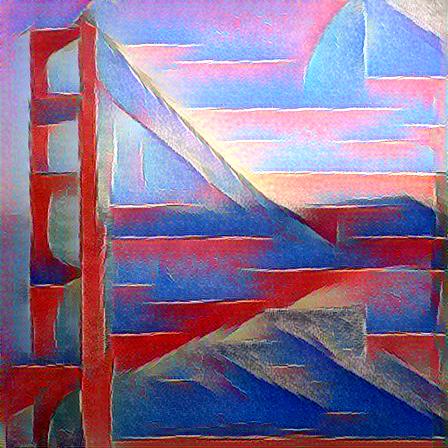}&
		\includegraphics[width=0.16\linewidth]{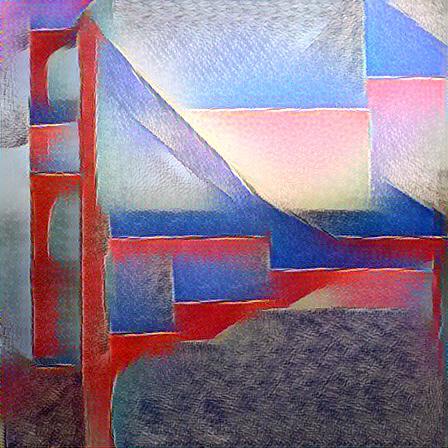}&
		\includegraphics[width=0.16\linewidth]{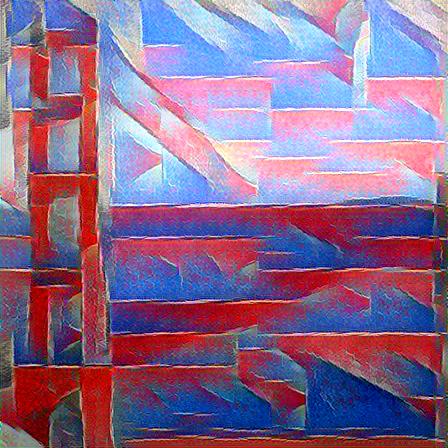}&
		\includegraphics[width=0.16\linewidth]{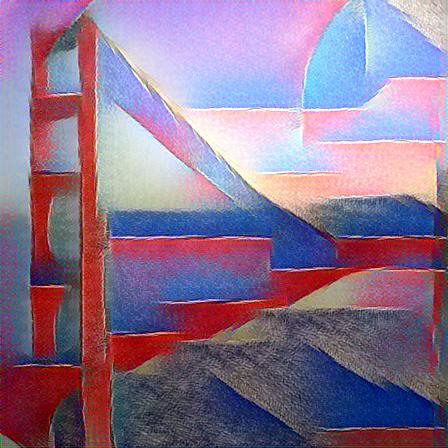}
		\\
		\scriptsize Style & \scriptsize CNNMRF &&&  \multicolumn{4}{c}{ \scriptsize CNNMRF + \bf  Our DivSwapper}
		
	\end{tabular}
	\caption{Our DivSwapper can diversify existing pure patch-based approaches like Style-Swap and CNNMRF, which is beyond the capability of SOTA diversified methods, e.g., DFP.} 
	\label{fig:quality2}
\end{figure}

{\bf Qualitative Comparison.} As shown in Fig.~\ref{fig:quality} (a,b), MTS and ITN only achieve subtle diversity, which is hard to perceive. In rows (c,e), DFP can diversify WCT and Avatar-Net to generate diverse results, but the diversity is still limited, especially for Avatar-Net. On the same baselines, our DivSwapper achieves much more significant diversity, e.g., the colors changed on the skies and buildings in rows (d,f) ({\em best compared with the difference heat maps in the last column}). Moreover, as shown in Fig.~\ref{fig:teaser} and~\ref{fig:quality2}, our DivSwapper can also diversify the pure patch-based methods like Style-Swap and CNNMRF, which is beyond the capability of DFP. It is worth noting that the patterns varied significantly in our results in Fig.~\ref{fig:quality2} generally correspond to the style regions with higher activation values in the activation heat maps, e.g., the beige walls in the top and the blue and red edges in the bottom. It verifies that our DivSwapper indeed helps vary more significant style patches with higher activation values. {\em We also validate its effectiveness on AdaIN~\cite{huang2017arbitrary} and SANet~\cite{park2019arbitrary} in SM}.

\renewcommand\arraystretch{0.5}

\newcommand{\tabincell}[2]{\begin{tabular}{@{}#1@{}}#2\end{tabular}}
\renewcommand\arraystretch{1}
\begin{table}[t!]
	\centering
	\setlength{\tabcolsep}{0.25cm}
	\begin{threeparttable}[b]
		\begin{tabular}{c|c|ccc}
			\hline
			\hline
			\scriptsize Baseline &\scriptsize Method&\scriptsize $D_{pixel}$& \scriptsize $D_{LPIPS}$ & \scriptsize Efficiency \\
			\hline
			\tabincell{c}{\scriptsize MTS} & \scriptsize Original &  \scriptsize 0.080 & \scriptsize 0.175 & - \\
			\hline
			\tabincell{c}{\scriptsize ITN} & \scriptsize Original &  \scriptsize 0.077 & \scriptsize 0.163 & - \\
			\hline
			
			\multirow{3}{*}[0in]{ \tabincell{c}{\scriptsize WCT}} & \scriptsize Original & \scriptsize 0.000&\scriptsize 0.000& \scriptsize 3.421s  \\
			&\scriptsize + DFP &  \scriptsize 0.162 & \scriptsize 0.431 & \scriptsize 4.091s \\
			&\bf \scriptsize + DivSwapper & \bf \scriptsize 0.204&\bf \scriptsize 0.485& \bf \scriptsize 3.566s \\
			\hline
			
			\multirow{3}{*}[0in]{ \tabincell{c}{\scriptsize Avatar-Net}} & \scriptsize Original & \scriptsize 0.000&\scriptsize 0.000& \scriptsize 3.920s  \\
			&\scriptsize + DFP & \scriptsize 0.102 & \scriptsize 0.264 & \scriptsize 4.268s \\
			&\bf \scriptsize + DivSwapper & \bf \scriptsize 0.128&\bf \scriptsize 0.320& \bf \scriptsize 3.932s \\
			\hline
			
			\multirow{2}{*}[0in]{ \tabincell{c}{\scriptsize Style-Swap}} & \scriptsize Original & \scriptsize 0.000&\scriptsize 0.000& \scriptsize 10.571s  \\
			&\bf \scriptsize + DivSwapper & \bf \scriptsize 0.065&\bf \scriptsize 0.234& \bf \scriptsize 10.582s \\
			\hline
			
			\multirow{2}{*}[0in]{ \tabincell{c}{\scriptsize \tnote{1}\, CNNMRF}} & \scriptsize Original & \scriptsize 0.084&\scriptsize 0.257& \scriptsize 118.44s  \\
			&\bf \scriptsize + DivSwapper & \bf \scriptsize 0.142&\bf \scriptsize 0.378& \bf\scriptsize 140.91s \\
			
			\hline
			\hline
		\end{tabular}
		\begin{tablenotes}
			\tiny
			\item[1] \scriptsize Due to the limitation of GPU memory, we only test the images of size 448$\times$448px for CNNMRF. 
		\end{tablenotes}
	\end{threeparttable}  
    \caption{Quantitative comparisons. The efficiency is tested on images of size $512\times 512$px and a 6GB Nvidia 1060 GPU. }
	\label{quantity}
\end{table}

{\bf Quantitative Comparison.} The quantitative results are shown in Tab.~\ref{quantity}. Consistent with Fig.~\ref{fig:quality}, MTS and ITN obtain low diversity scores in both Pixel and LPIPS distance. Integrated into the same baselines (i.e., WCT and Avatar-Net), our DivSwapper is clearly superior to DFP in both diversity and efficiency (DFP involves some slow CPU-based SVD operations to obtain orthogonal noise matrix). In addition, our DivSwapper can also diversify Style-Swap and help improve the diversity of CNNMRF. Note that due to the use of noise initialization and iterative optimization process, CNNMRF has produced some varied results and the extra time increased by our DivSwapper is more than other baselines.

{\bf Quality Comparison.} As style transfer is highly subjective, we conduct a user study to evaluate how users may prefer the outputs of our diversified methods over the deterministic ones and those of other SOTA diversified methods (i.e., DFP). WCT and Avatar-Net are adopted as the baselines. Twenty users unconnected with the project are recruited. For each baseline, we give each user 50 groups of images (each group contains the input content and style images, and three randomly shuffled outputs, i.e., one original output of the baseline method, one random output of baseline + DFP, and one random output of baseline + our DivSwapper) and ask him/her to select the favorite output. The statistics in Tab.~\ref{userstudy} show that both DFP and our DivSwapper can help users obtain preferred (higher quality) results compared with baselines, and our method also achieves higher quality than DFP.

\renewcommand\arraystretch{1}
\begin{table}[t!]
	\centering
	\setlength{\tabcolsep}{0.4cm}
	\begin{center}
		\begin{tabular}{cccc}
			\hline
			\hline
			\scriptsize Baseline & \scriptsize Original & \scriptsize + DFP & \scriptsize + Our DivSwapper
			\\
			\hline
			\scriptsize WCT & \scriptsize 27.24 & \scriptsize 33.96 & \scriptsize \bf 38.80
			
			\\
			\scriptsize Avatar-Net & \scriptsize 28.81 & \scriptsize 31.78 & \scriptsize \bf 39.41
			\\
			\hline
			\hline	
		\end{tabular}
	\end{center}
    \caption{Percentage (\%) of the votes in the user study.}
	\label{userstudy}
\end{table}

\renewcommand\arraystretch{0.9}
\begin{figure}[t!]
	\centering
	\setlength{\tabcolsep}{0.01cm}
	\begin{tabular}{cccccp{0.04cm}|p{0.04cm}c}
		\includegraphics[width=0.16\linewidth]{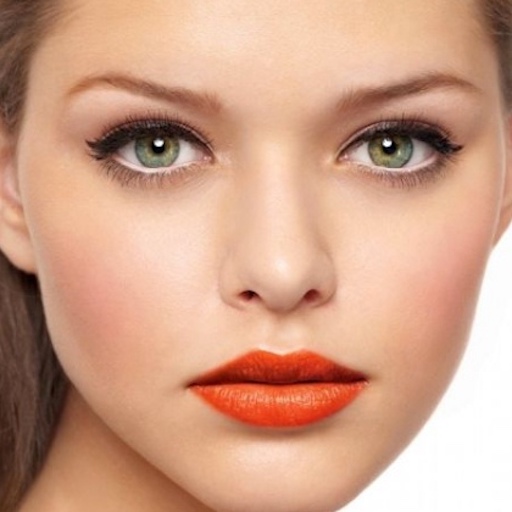}&
		\includegraphics[width=0.16\linewidth]{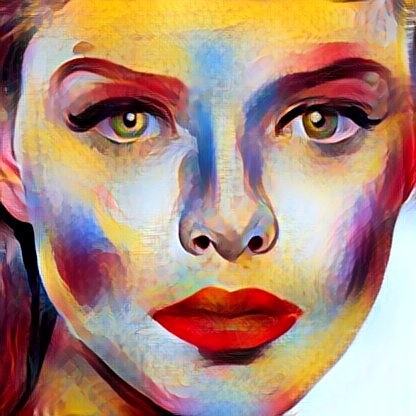}&
		\includegraphics[width=0.16\linewidth]{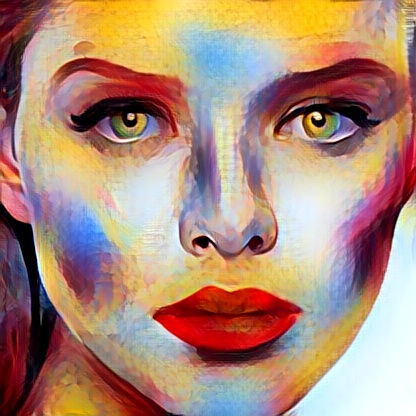}&
		\includegraphics[width=0.16\linewidth]{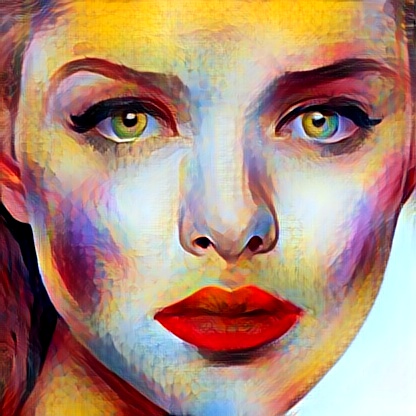}&
		\includegraphics[width=0.16\linewidth]{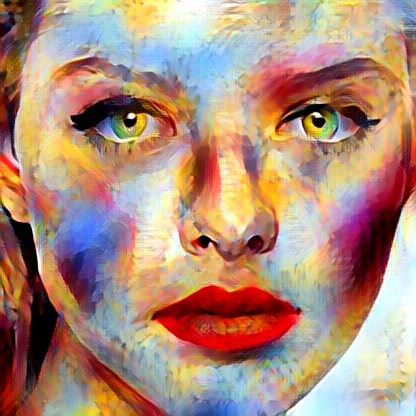}&
		&&
		\includegraphics[width=0.16\linewidth]{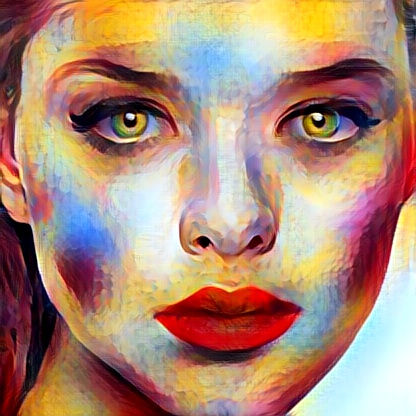}
		\\
		
		\includegraphics[width=0.16\linewidth]{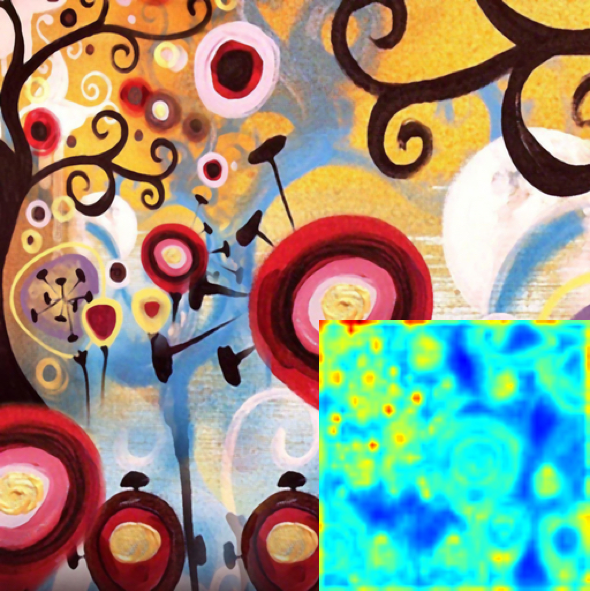}&
		\includegraphics[width=0.16\linewidth]{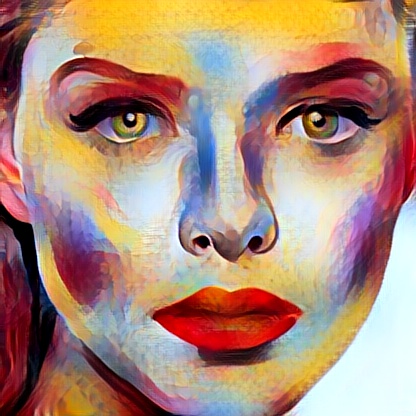}&
		\includegraphics[width=0.16\linewidth]{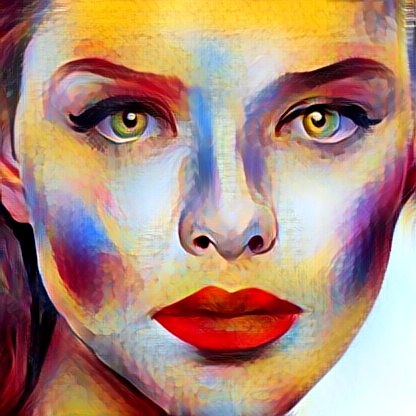}&
		\includegraphics[width=0.16\linewidth]{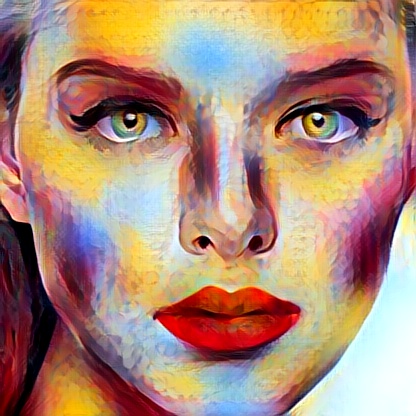}&
		\includegraphics[width=0.16\linewidth]{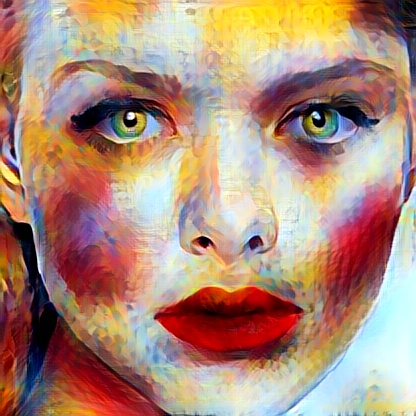}&
		&&
		\includegraphics[width=0.16\linewidth]{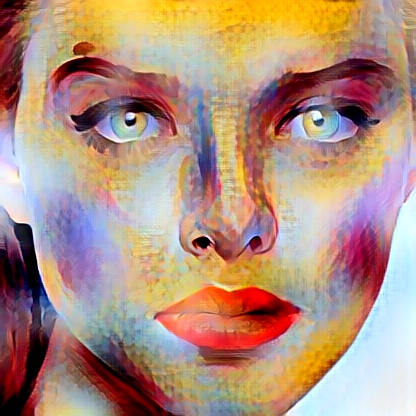}
		\\
		\scriptsize Inputs &\tiny (0, 5$\times10^1$] &\tiny (0, 5$\times10^2$] &\tiny \bf {(0, 5${\bf\times10^3}$]} &\tiny (0, 5$\times10^4$]  &
		&& \scriptsize Normal
		\\
		\hline
		\scriptsize $D_{pixel}$ & \scriptsize 0.043 & \scriptsize  0.077 &  \scriptsize \bf 0.128 &  \scriptsize 0.145 && &\scriptsize 0.124
		
		\\
		\hline
		\scriptsize $D_{LPIPS}$ & \scriptsize 0.085 &  \scriptsize 0.187 &  \scriptsize \bf 0.320 &  \scriptsize 0.389 && &\scriptsize 0.311
	\end{tabular}
	\caption{Effects of different sampling ranges ($2^{nd}$ to $5^{th}$ columns) and distributions (last column) of the random deviations $\sigma$. Our DivSwapper is integrated into Avatar-Net~\protect\cite{sheng2018avatar}.}
	
	\label{fig:tradeoff}
\end{figure}

\subsection{Ablation Study}
\label{abs}

{\bf Graceful Control between Diversity and Quality.} Our DivSwapper can provide graceful control between diversity and quality by sampling the deviations $\sigma$ from different ranges. As shown in Fig.~\ref{fig:tradeoff}, with the increase of sampling range, the generated results consistently gain more diversity (which can also be validated by the metrics below), but a too-large sampling range may reduce the quality (e.g., the $5^{th}$ column). When proper range (e.g., $(0, 5\times10^3$]) is applied, we can obtain the sweet spot of the two: the results exhibit considerable diversity and also maintain the original quality and some inherent characteristics. For different baselines, the proper range of $\sigma$ can be easily determined via only a few trials and errors, and our experiments verify that {\em these constant range values can work stably on different content and style inputs}.

{\bf Effect of Sampling Distribution.} We also try other sampling distributions instead of the default uniform one. As shown in the last column of Fig.~\ref{fig:tradeoff}, sampling $\sigma$ from a normal distribution could achieve similar performance (e.g., the top image and the diversity scores), but the results may be erratic and sometimes produce unwanted effects (e.g., the hazy blocks in the bottom image). This problem may be caused by the concentration property of the normal distribution. However, it does not occur when using a uniform distribution.

\begin{figure}[t!]
	\centering
	\setlength{\tabcolsep}{0.015cm}
	\begin{tabular}{ccccp{0.01cm}|p{0.01cm}cc}
		\includegraphics[width=0.16\linewidth]{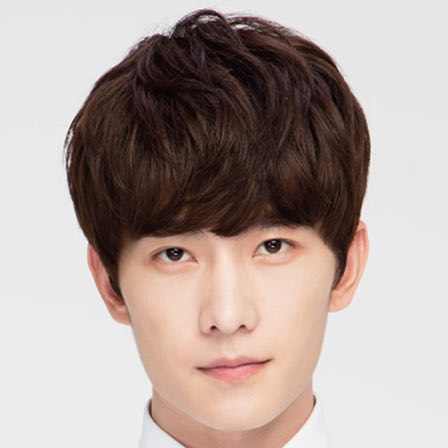}&
		\includegraphics[width=0.16\linewidth]{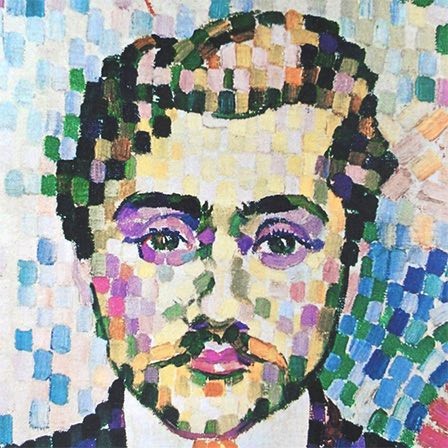}&
		\includegraphics[width=0.16\linewidth]{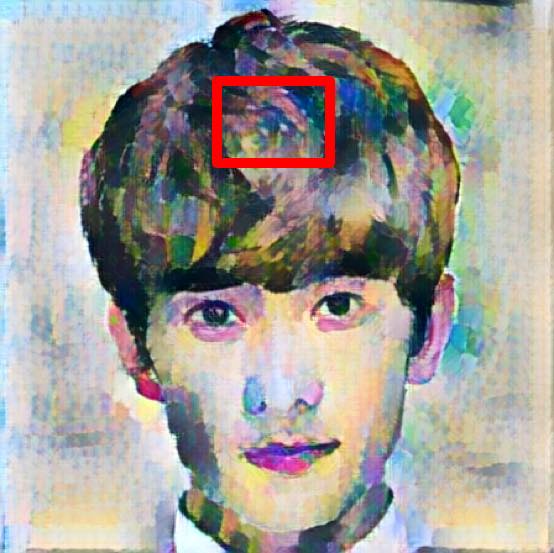}&
		\includegraphics[width=0.16\linewidth]{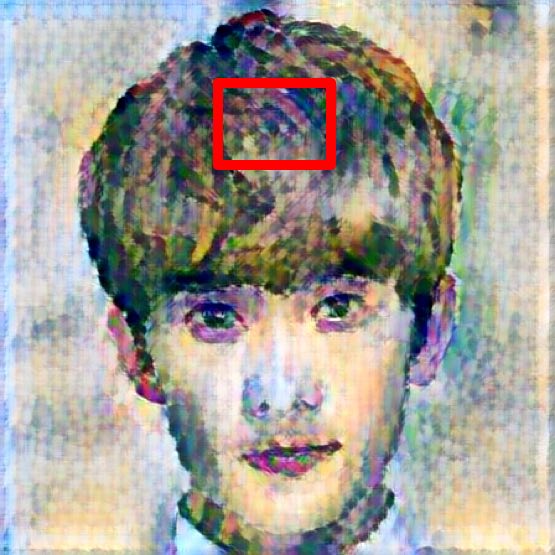}&&&
		\includegraphics[width=0.16\linewidth]{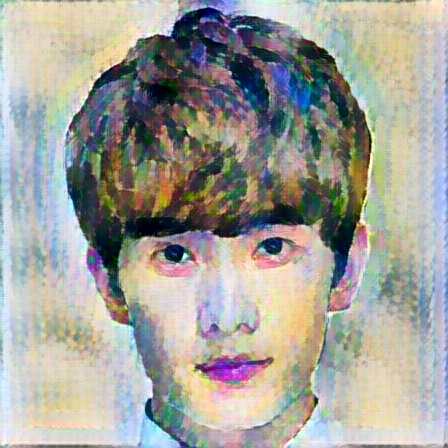}&
		\includegraphics[width=0.16\linewidth]{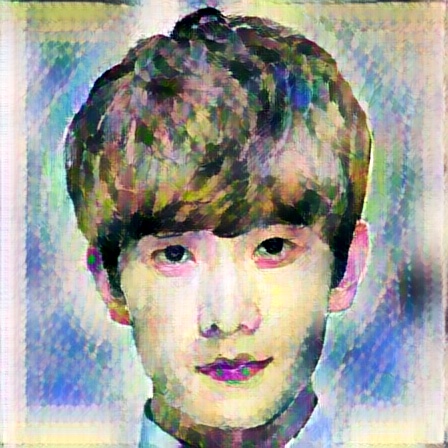}
		\\
		
		\includegraphics[width=0.16\linewidth]{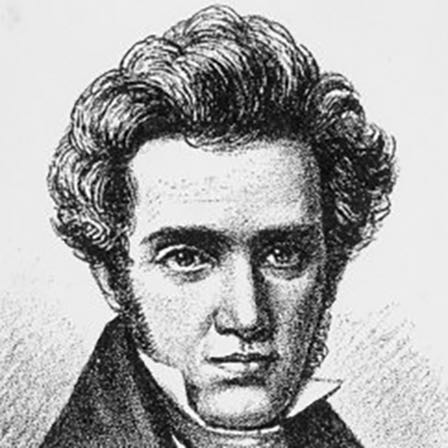}&
		\includegraphics[width=0.16\linewidth]{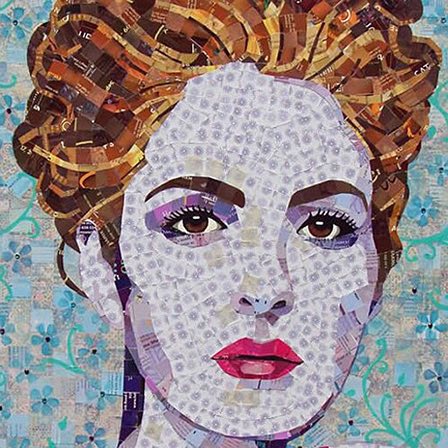}&
		\includegraphics[width=0.16\linewidth]{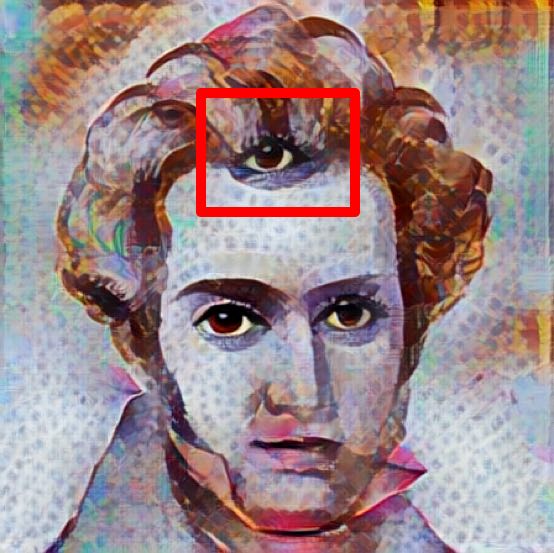}&
		\includegraphics[width=0.16\linewidth]{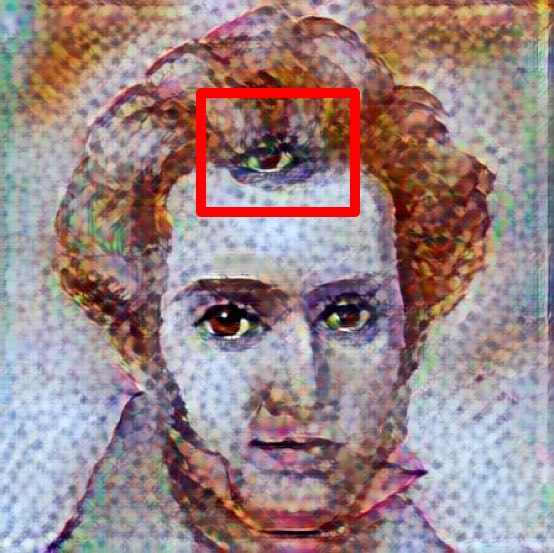}&&&
		\includegraphics[width=0.16\linewidth]{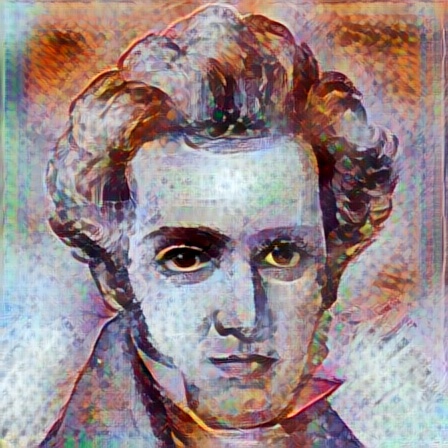}&
		\includegraphics[width=0.16\linewidth]{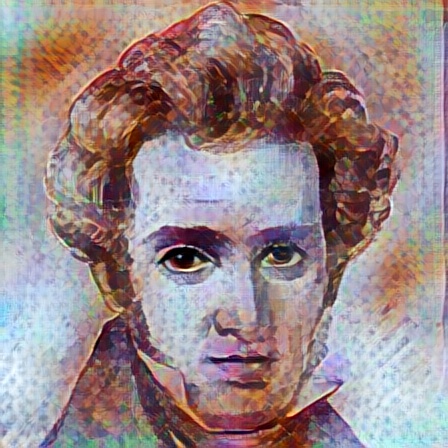}
		\\
		
		\scriptsize (a) Content & \scriptsize (b) Style & \scriptsize (c) Avatar-Net & \scriptsize (d) + DFP  &&& \multicolumn{2}{c}{\scriptsize (e) + Our DivSwapper}
		
	\end{tabular}
	\caption{ Results on semantic-level style transfer, e.g., portrait-to-portrait. The diverse results generated by using our DivSwapper can maintain the semantic-level stylization while also alleviating the undesirable artifacts (e.g., the eye patterns in the red box areas) caused by the restricted patch matching.} 
	\label{fig:portraits}
\end{figure}

{\bf Semantic-level Style Transfer.} Though our primary motivation is to improve the diversity in {\em artistic style transfer}, for semantic-level style transfer, the proposed method can also produce diverse results while maintaining the original quality and the semantic correspondence. As can be seen in our generated results in Fig.~\ref{fig:portraits} (e), although the patterns in each semantic area vary significantly (e.g., the backgrounds and hairs), the main semantic-level stylization is well preserved (e.g., the details on portraits). It is because the SSN used in our DivSwapper is still constrained by the original NCC, and the variations are gracefully controlled by $\sigma$, as analyzed in Sec.~\ref{method}. Moreover, our DivSwapper can also help alleviate the inherent flaws (e.g., undesirable artifacts) caused by the original restricted patch matching~\cite{zhang2019multimodal}, which is our new merit against SOTA methods, e.g., DFP in column~(d). It also further justifies that our method can help users obtain diverse results with higher quality.

\renewcommand\arraystretch{0.5}

\section {Concluding Remarks}
\label{conclude}
In this work, we explore the challenging problem of diversified patch-based style transfer and introduce a universal and efficient module, i.e., {\em DivSwapper}, to resolve it. Our DivSwapper is plug-and-play and can be easily integrated into existing patch-based and Gram-based methods to generate diverse results for arbitrary styles. Theoretical analyses and extensive experiments demonstrate the effectiveness of our method, and compared with SOTA algorithms, it shows superiority in diversity, quality, and efficiency. We hope our analyses and investigated method can help readers better understand the crux of patch-based methods and inspire future works in style transfer and many other similar fields. 

\section*{Acknowledgements}
This work was supported in part by the projects No. 2021YFF0900604, 19ZDA197, LY21F020005, 2021009, 2019011, Zhejiang Elite Program project: research and application of media fusion digital intelligence service platform based on multimodal data, MOE Frontier Science Center for Brain Science \& Brain-Machine Integration (Zhejiang University), National Natural Science Foundation of China (62172365), and Key Scientific Research Base for Digital Conservation of Cave Temples (Zhejiang University), State Administration for Cultural Heritage.


{\small
   \bibliographystyle{named}
   \bibliography{ijcai22}
}

\end{document}